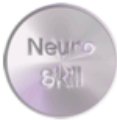

# NeuroSkill™: Proactive Real-Time Agentic System Capable of Modeling Human State of Mind

**Nataliya Kosmyna**[1]
*MIT Media Lab*
*Cambridge, MA*

**Eugene Hauptmann**
*MIT*
*Cambridge, MA*

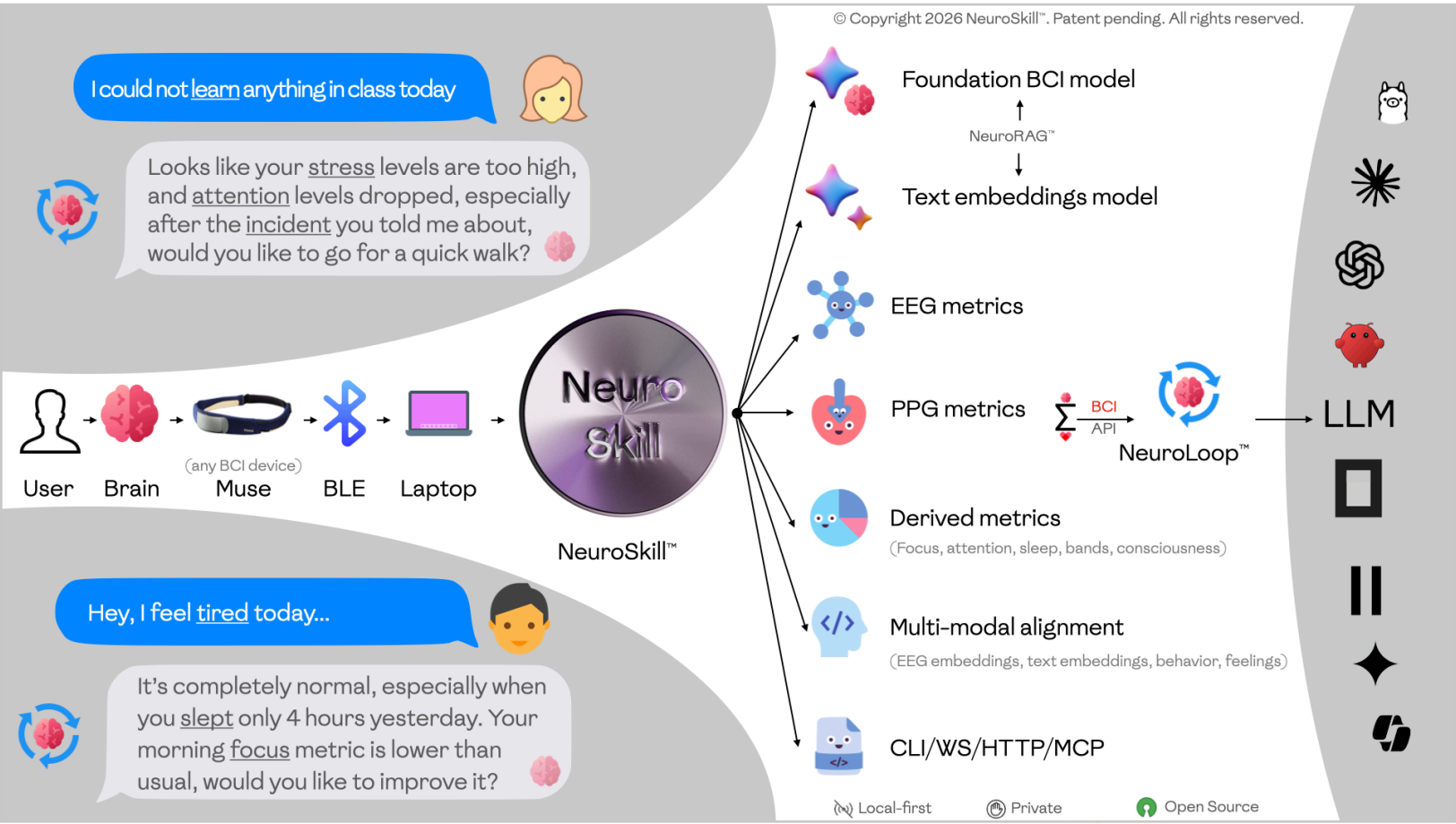

Figure 1. Overall real-time neuroadaptive agentic harness system architecture.

## Abstract

Real-time proactive agentic system, capable of modeling Human State of Mind, using foundation EXG model and text embeddings model, running fully offline on the edge. Unlike all previously known systems, the NeuroSkill™ system leverages SKILL.md description of Human's State of Mind via API and CLI provided by the system, directly from the Brain-Computer Interface (BCI) devices, which records Human biophysical and brain signals. Our custom harness – NeuroLoop™ – utilizes all of the above to run agentic flow that manages to engage with the Human on multiple cognitive and affective levels of their State of Mind (e.g., empathy), by providing actionable tool calls and protocol execution with explicit or implicit requests from the Human. GPLv3 open-source software with ethically aligned AI100 licensing for the skill markdown.

[1] Nataliya Kosmyna is the corresponding author, please contact her at nkosmyna@mit.edu

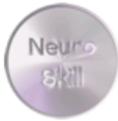

*"My wife has always been eager to change the world. But I'll just settle for understanding it first."*

– Will Caster. Transcendence. 2014

## TL;DR

Download system from neuroskill.com, install it and connect to your BCI device, then go to neuroloop.io and install the harness `npm install -g neuroloop` and connect it to your LLM of choice (ollama via /model command or /login command to connect to commercial ones). If you want to quickly understand how it works, start with the system architecture.

## Open Source

Proposed solution is open sourced under GPLv3 "copyleft" license, with the markdown files available on AI100 license.

- NeuroSkill™ app source code – https://github.com/NeuroSkill-com/skill
- NeuroLoop™ LLM harness – https://github.com/NeuroSkill-com/neuroloop
- NeuroSkill™ markdown files – https://github.com/NeuroSkill-com/skills
- Command Line Interface (CLI) tool – https://github.com/NeuroSkill-com/neuroskill

- Website of the main app – https://neuroskill.com
- Website of the LLM harness – https://neuroloop.io
- Website for the CLI – https://neuroskill.dev

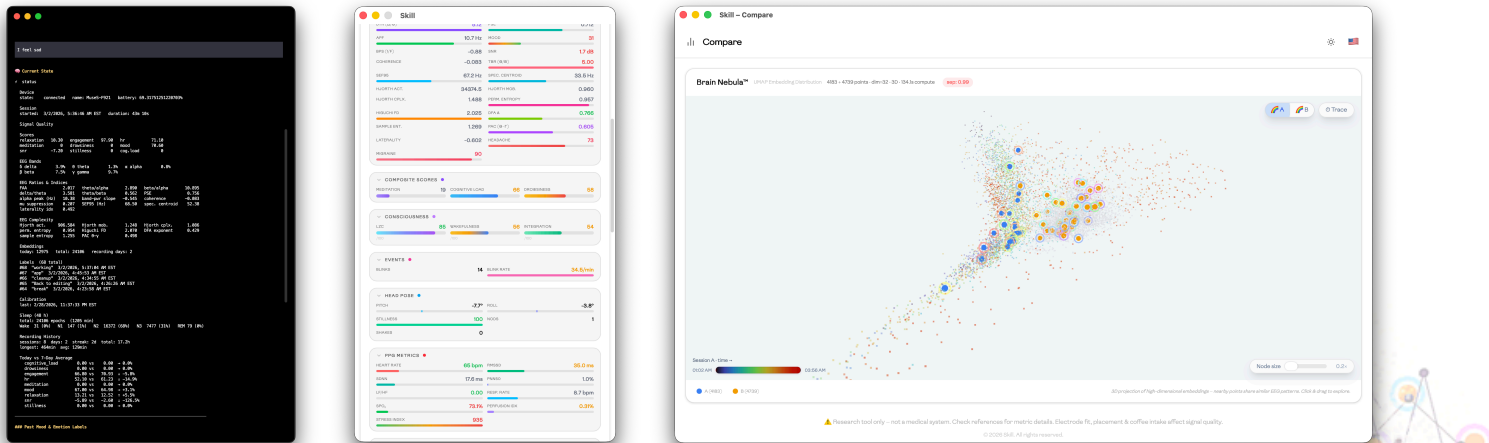

Figure 2. From left to right. NeuroLoop™ LLM harness interface. NeuroSkill™ main UI interface. Brain Nebula™ searchable interface embeddings of the Human's State of Mind.

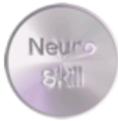

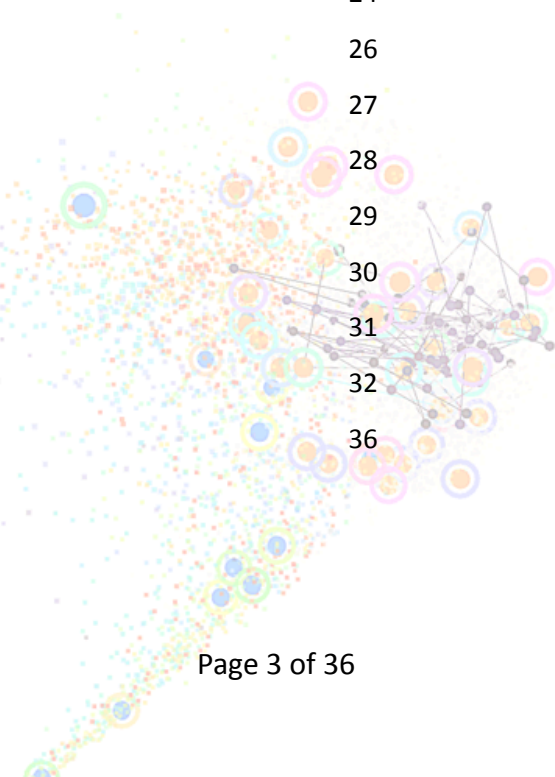

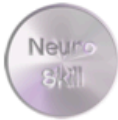

## Introduction

According to a report from Menlo Ventures [1] there were about 1.7-1.8 billion people who used large language model (LLM)-based systems, with about 600 million people who use LLMs daily.

Recent success of autonomous agents showed that there is demand for LLM harnesses that are unprecedented in their intent and their agency.

While systems such as OpenClaw [2] aim to provide autonomy to agents, this work explores a different dimension of that relationship: **how an agent can understand what a Human feels, what a Human does, and how a Human's experiences shape their behavior – ultimately influencing the interaction and communication between the agentic system and the Human.**

In this work we present a novel system that is designed to build foundational representation of the brain states provided by the Human for the Agent to index, align, search, and navigate. In most cases it is done in real-time and on the edge (your local personal computer), while engaging with the Human; in other cases – to perform deep sleep research while Human is not in an active conscious state.

We offer a simple set of markdown files backed by the Application Programming Interface (API) and Command Line Interface (CLI) interfaces we released, using open source license GPLv3, that allows the LLM harness to execute the agent, and allows agent to engage with a new set of biophysical data that carefully models Human State of Mind, behavior, emotions, social cues, relationships, and anything the user can put in writing, via voice, or even when the user does not have to or cannot do it: they might just feel it, think it, imagine it, or experience it.

We call this system **NeuroSkill**™.

It encapsulates the app that always runs and collects data from the Brain-Computer Interface (BCI)-enabled wearable, non-invasive devices (the system can be extended to ingest data from any invasive and noninvasive device available on the market today). The access is provided via an API *npx neuroskill <command>*, and can be edited as described in the published markdown SKILL.md files.

The LLM harness we designed is called **NeuroLoop**™.

It is a unique implementation, because it relies not only on the tools and skills provided to the system, but it parses in real-time whether additional data needs to be requested, aligned, or generated. As a result, it proactively adjusts its recommendations for the Human behavior to align with the goals the Human initially sets up as the Agent's main objective.

The proposed system is highly adjustable and modifiable. A Human with no coding skills can easily extend it by merely providing new markdown files and explaining in plain English how they want the agent to use them inside the NeuroLoop™.

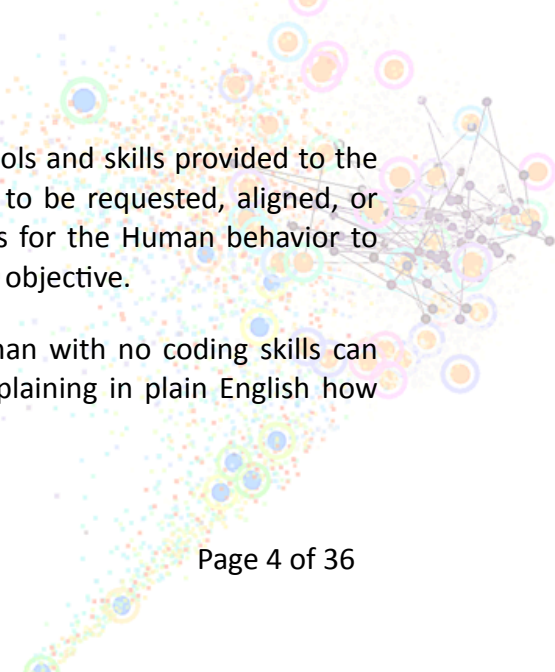

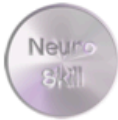

## Motivation

Emerging research raises concerns about the cognitive and societal implications of LLM usage [25]. The inception of agentic systems offers more flexible interfaces than expensive LLM fine-tuning and greater versatility than Retrieval-Augmented Generation (RAG) systems.

Mass adoption of wearable devices continues to increase year over year. Affordable EXG systems like Muse [26], OpenBCI [27] or AttentivU [28] make consumer-grade BCI devices ubiquitous by design.

Big tech corporations collect vast datasets from users of their devices such as the Apple Watch, Samsung Galaxy Watch, Fitbit, Oura, and others. Very few of these datasets are published or actively used to benefit the very individuals from whom they were collected.

With brain data, the risk–benefit model is different, and as a species, we cannot afford to let BCI innovation be concentrated in the hands of a few.

Our goal in publishing this work is to encourage neuroscientists, developers, caregivers, students, and others to view the value of their brains differently and to make their own brains work for them.

We open-source the full implementation of the system so **anyone can use it**.

Even better, the NeuroLoop™ system itself can help its user modify it and its source code, making it highly tailored to the user's needs — all **without violating the user's privacy, rights, or dignity**.

Recognizing that this is the first system of its kind — built on decades of scientific and open-source effort by humanity — we welcome feedback and aim to contribute to a better future for Brain-Computer Interfaces.

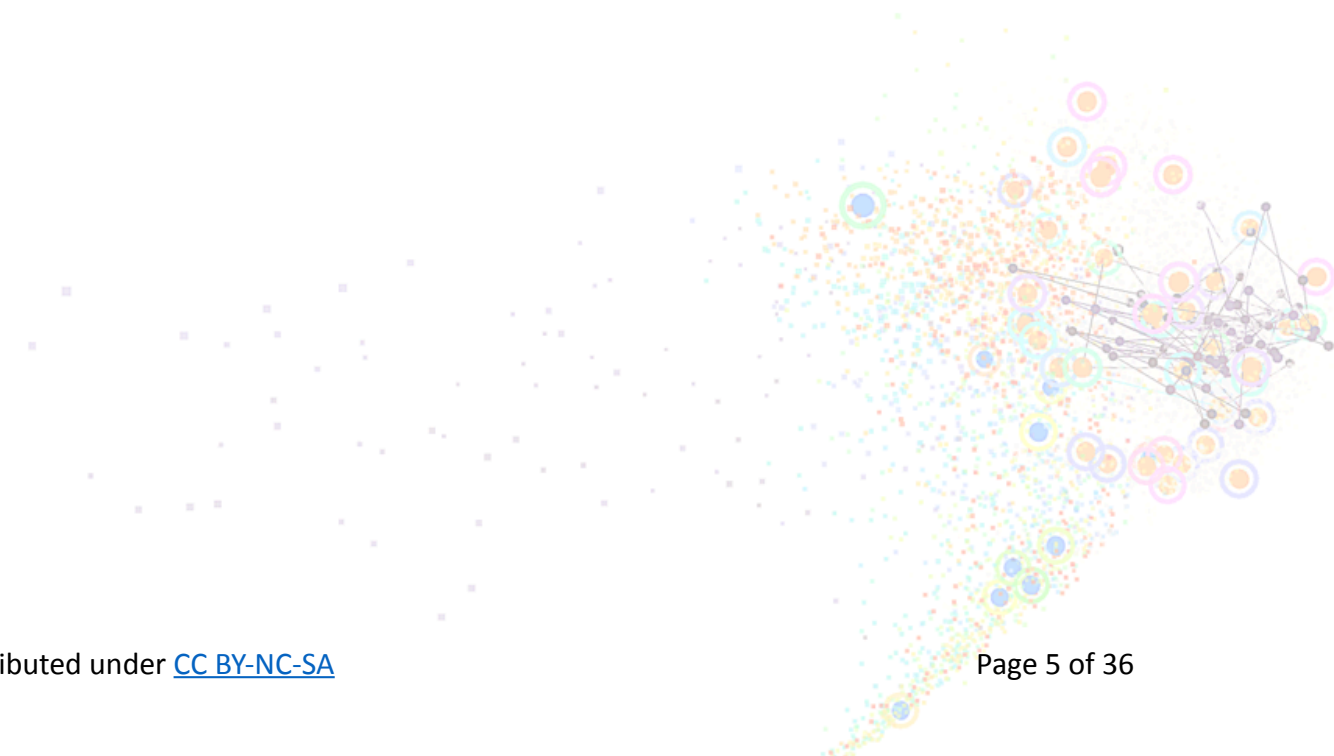

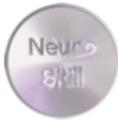

## Background and Related Work

### Cognitive and Affective States of the User

Generative AI is reshaping all spheres of one's life by enabling personalized, on-demand experiences in learning, personal relationship advices, work, health, legal and much more. However, current agentic systems lack awareness of the user's cognitive and affective states, limiting their adaptability.

Current frameworks and guidelines assume a Human as a user, and user-only, with a set of static needs and parameters: a rational actor who continuously processes information in an objective manner, understands the functioning of the framework at all times, maintains stable judgment under conditions of novelty and uncertainty and has a stable, predictable reaction and response to outputs of the chatbot or an agent. In most cases, the current agentic frameworks do not consider if/when Human can have any type of strong reaction, while interacting with the bot, not only due to the responses and output obtained, but also considering any other, external factors, that might influence the judgement, reaction times, affective and cognitive responses of the user. Most of current implementations are built on the fact that the user may inform the system of their state and other relevant information, but they might not consider doing so, as they may not realize that this information could improve the agent's output. A user may also feel uncomfortable sharing the details of the event that influenced their state, yet may be willing to indicate how they feel overall. Finally, it can be difficult for a user to accurately assess their own emotions, making an objective measure of affective and cognitive states potentially valuable for the agent.

Consider this use case: a user is currently stuck at Boston Logan Airport due to inclement weather and is about to miss a jubilee celebration for their parents' wedding anniversary. The airline's agents are unable to help rebook the user on their preferred flight, offering only a departure 72 hours after the original schedule. The user might feel anger, anxiety, and sadness due to this situation, however, the standard "take a deep breath" or "please calm yourself" responses from an LLM might not help in this situation, as they could frustrate the user even further. The user might benefit from a simple "I am available if you need me" response, where an "inaction" from the agent might be more beneficial to the user. Currently, the **agent does not know** and has no way of measuring or accessing the user's reaction to its suggestions in order to adapt accordingly. If the agent can map their suggestions and the user's reactions to them, it can build the State of Mind over time, learning from previous similar stressful situations. The agent will have information indicating that suggesting a user calm down at a certain level of stress and anger can escalate their frustration.

A truly adaptive agentic system should not only be capable of generating adaptive content but also recognize cognitive and affective states of the user and taking those into consideration – an ability that is fundamental to effective Human communication and collaboration.

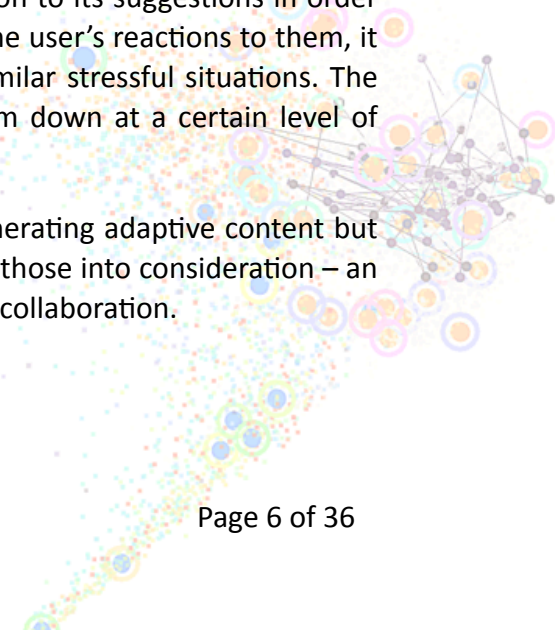

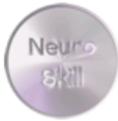

### Neuroadaptive AI interfaces

The integration of generative AI with BCIs represents an emerging and rapidly evolving research domain. Although machine learning techniques have long been used for the analysis and decoding of neural signals, the majority of AI-enhanced BCI systems have focused on brain state classification and signal reconstruction, rather than on interactive, real-time adaptive use cases. Generative AI introduces new conceptual and technical possibilities, extending BCIs beyond passive neural decoding, enabling dynamic content modulation and interactive adaptation. Preliminary studies suggest that the integration of LLMs within BCI frameworks may advance Human–Computer Interaction (HCI), with potential applications spanning assistive technologies for communication, education as well as cognitive augmentation [3, 4].

### Using Generative AI to Analyze Brain Data

A major focus in AI-(non-invasive)-BCI research has been EEG-based brain decoding, where generative AI and machine learning models are used to encode and decode the neural signals underlying visual or auditory information processing [5-7]. While these methods advance neural signal processing, they remain limited in real-time user interaction. Readers interested in these approaches can refer to a comprehensive review by Sabharwal and Rama (2024) [8]. Beyond decoding, LLMs have been applied to EEG for brain state classification and assistive communication [9]. In clinical applications, for example, language model-enhanced BCI communication systems have improved typing accuracy for ALS patients by up to 84% in real-time BCI spelling sessions [10]. Subsequent approaches have demonstrated that LLMs can classify brain states at the word level from EEG data during reading tasks [11, 12].

Recent research has extended these applications to foundation models that generalize across EEG tasks, such as NeuroLM [13], Neuro-GPT [14], Zuna [15], REVE [16], Luna [17] among others. Some of these models are trained predominantly (or exclusively) on clinical recordings, and thus they may underperform on EEG from neurologically healthy or other non-clinical participants, highlighting the need for cross-domain evaluation or adaptation when deploying these models. We refer the reader to a comprehensive review by Liu et al. (2026) [18] to learn more about existing EEG foundation models.

Other approaches have explored personal health and well-being. For example, Sano et al. (2024) [19] used LLMs to interpret EEG signals for sleep quality assessment, providing tailored recommendations. Similarly, EEG Emotion Copilot [20] integrated EEG with an LLM to analyze EEG signals, identify affective states, and generate automated clinical insights. MultiEEG-GPT [21] integrates EEG with facial expressions and audio to enhance mental health assessments using LLM-based classification.

Finally, generative AI techniques have been leveraged for data augmentation to enhance EEG-based model training [22, 23]. However, these approaches mainly focus on recognizing states and are yet to support real-time user interaction.

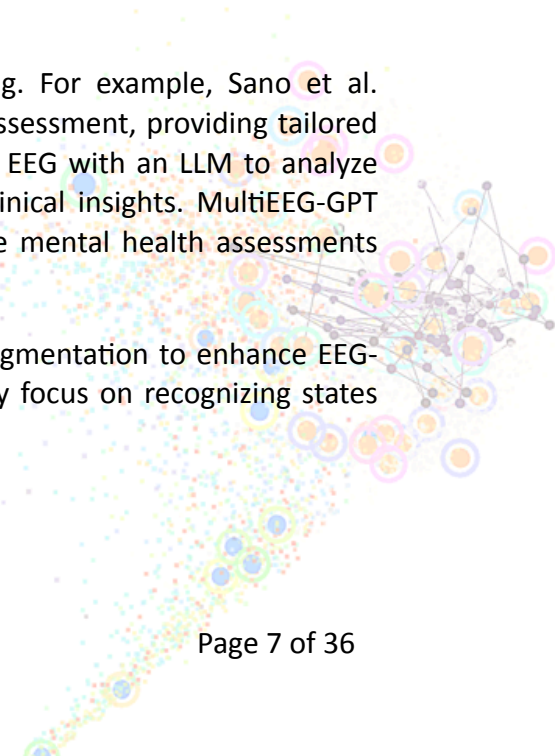

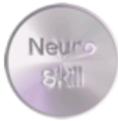

### Neuroadaptive Generative AI Systems

One of the few neuroadaptive systems integrating generative AI for real-time adaptation is AdaptiveCoPilot [24], designed for expert pilots in virtual reality (VR). AdaptiveCoPilot continuously adjusts visual, auditory, and textual cues based on real-time cognitive load assessments, optimizing performance in high-stakes environments. It relies on functional near-infrared spectroscopy (fNIRS) and is tailored for high-performance cognitive tasks. Finally, NeuroChat [4], one of the first systems to integrate EEG-based cognitive state tracking (engagement) with generative AI in real-time for learning use cases.

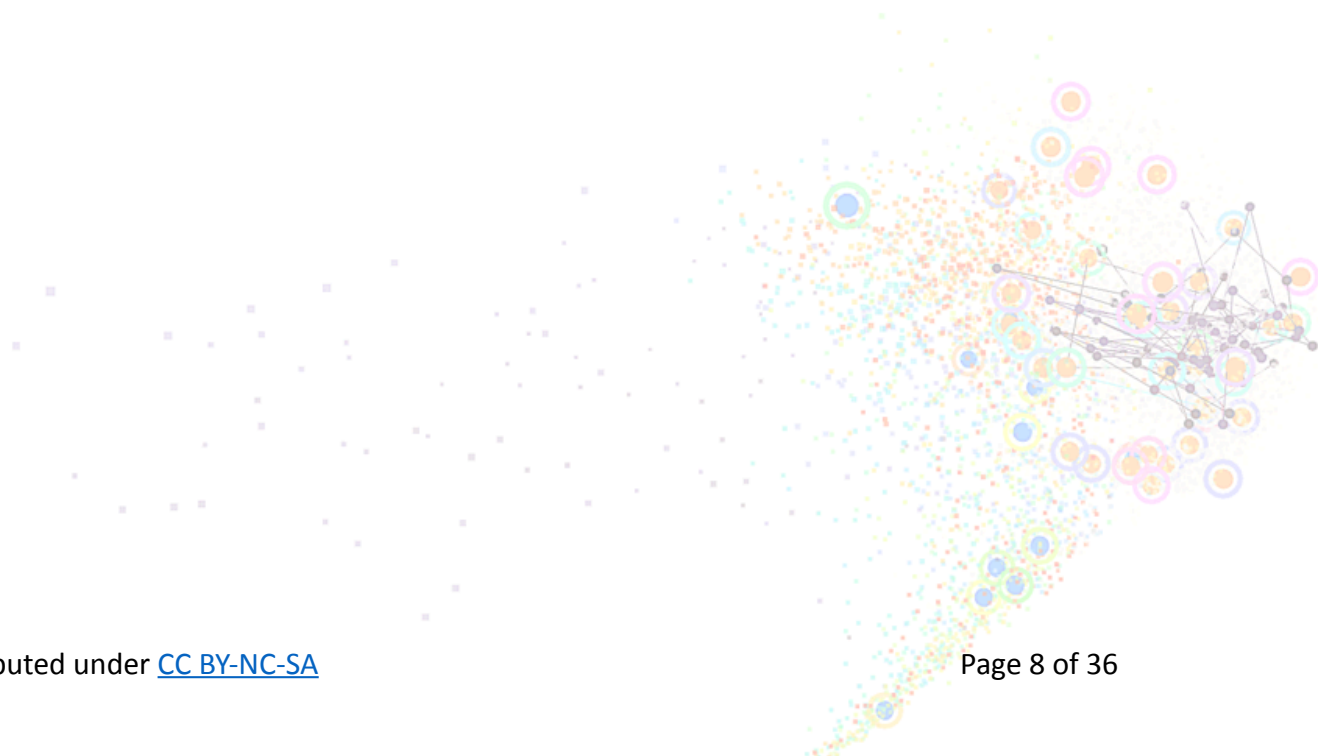

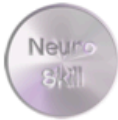

## Use Cases

Consider a diverse range of individuals — students, busy parents, pilots, doctors, office managers, delivery drivers, knowledge workers, stay-at-home parents, retirees, neurodivergent individuals, and those with disabilities — all of whom could engage with agents in new ways. While many already use LLMs on a daily basis, only a few have begun to adopt an agentic approach. OpenClaw [2] has played a role in igniting some interest from the public. However, as is often the case, most of these interactions are still primarily conducted in natural language, such as English.

Not all individuals are able to freely express themselves; at times they may be unable, unwilling, or unsure how to communicate their thoughts and feelings. Moreover, certain experiences may transcend the limitations of language.

To achieve this goal, we designed the system to intake the State of Mind of the Human and expose it to the Agent in a searchable, navigable, and explainable form. Even when language descriptions are not attached to the state, the NeuroLoop™ and NeuroSkill™ system can project the underlying state, perform comparisons, and assist Human in navigating the information to achieve their objectives.

## Education

Let's imagine an educational use case, a popular scenario, covered in multiple AI learning papers: a student who is overworked from trying to manage the workload of the midterms, and tries to study hard, but simply cannot catch up. In this case a student might do a study session and engage a Focus Protocol, and briefly interact with the NeuroLoop™ by asking questions about their subject. The Agent would know how engaged and tired the student is, if it needs to simplify the answer for the student, or engage in the study protocol, ask follow-up questions, engage the student on the metacognitive topics on how to learn something, or when the student's State of Mind has a trend where it cannot ingest more information effectively. The NeuroLoop™ system monitors the student's cognitive state to provide timely interventions, such as suggesting a short break, a stretch, or a quick snack. If a student becomes distracted — whether by a phone call, gaming, or a social media "doom scroll" session — the system detects these shifting dynamics and implements a recovery protocol designed to re-engage their focus on a task at hand. It is important to note that this system is designed to allow students to make choices on their own behalf. Using the system through parental controls grants parents excessive authority, effectively depriving students of their autonomy. This imbalance can undermine the student's sense of autonomy and negatively impact their Human dignity, which goes against ethics statement of this work and authors' values.

The NeuroLoop™ system can be proactive and can be used in monitoring state, all of it is configurable via simple markdown system outlined in the NeuroSkill™ architecture. It can engage the university- or school-provided LLM, which has access to a curriculum-based RAG system or internal search index authorized by the school, parents, state, or other relevant authorities. Because the NeuroLoop™ system is flexible - it can include additional tool and function calls, adopt SOUL.md identities, or introduce additional skills it might need to fulfill its Human-user's objective. Together with the state management, memory, code to navigate, parse and make decisions about the biophysical State of Mind and brain data, this internal design can be extended to the skills that are provided outside the NeuroSkill™ markdown system as well.

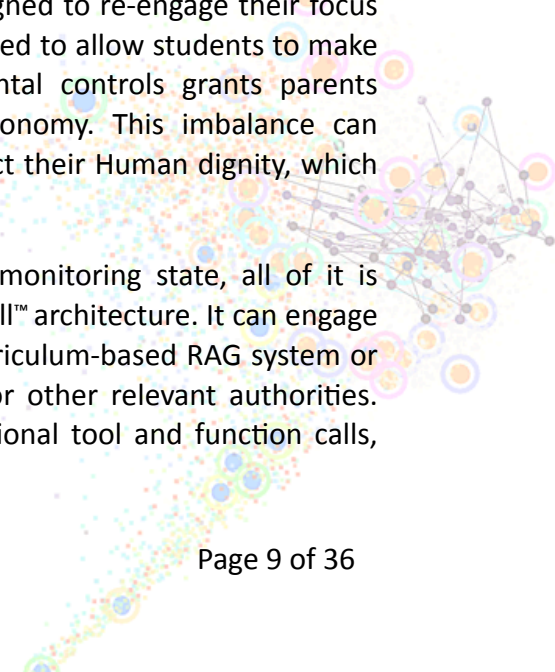

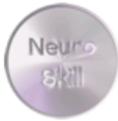

### Gaming

Now let's consider a different use case – gaming.

People play games for a variety of reasons, including social interaction, relaxation, learning, competition, storytelling, and leisure. Most players seek different benefits from their gaming experience.

Modern gaming is highly engaging, and many games are social — ranging from platforms like Roblox [30] to Massively Multiplayer Online Role-Playing Games (MMORPGs) such as World of Warcraft [31]. Some games can be stress-inducing, like Fortnite [32]. The list of game experiences is rich: competitive, relaxing, challenging, immersive, strategic, cooperative, emotional, educational, creative, adventurous, and suspenseful.

Nevertheless, the Human engages with games for their own reasons, and can experience a wide range of states — entering flow state [29], or feeling agitated, stressed, scared, sad, overwhelmed, as well as happy, relaxed, hyperactive, confused, or lost.

Managing such states can be complicated, especially for children and young adults, and remains challenging for adults navigating work and social environments. The healthcare and wellness industries frequently use games to help patients achieve specific mental or emotional states or practice certain skills. Educators may leverage game formats to help students better comprehend the curriculum and stay engaged. Parents might use games as a reward system. Businesses often employ gamification techniques to influence dopamine and stress-related hormone responses, for example by leveraging fear of missing out (FOMO) or social media doom-scrolling.

Such guided experiences can both help and harm the player, but most of the time the player is subjected to external factors and may not be aware of their influence on their State of Mind and body.

Systems like NeuroSkill™ can be of help: from implicit passive state tracking – to active intervention with a prescribed or recommended protocol based on the NeuroLoop™ values and objectives defined by its Human user. Similar to Parental Control or Screen Time features, the NeuroLoop™ system can help manage time spent playing the games, or manage and limit its influence on the wellbeing of the Human, without sacrificing the "good" parts prioritized by the Human via the explicit or implicit signals such as their brain data or markdown file describing how their biophysical and emotional states should be correlated based on the gaming experience they have.

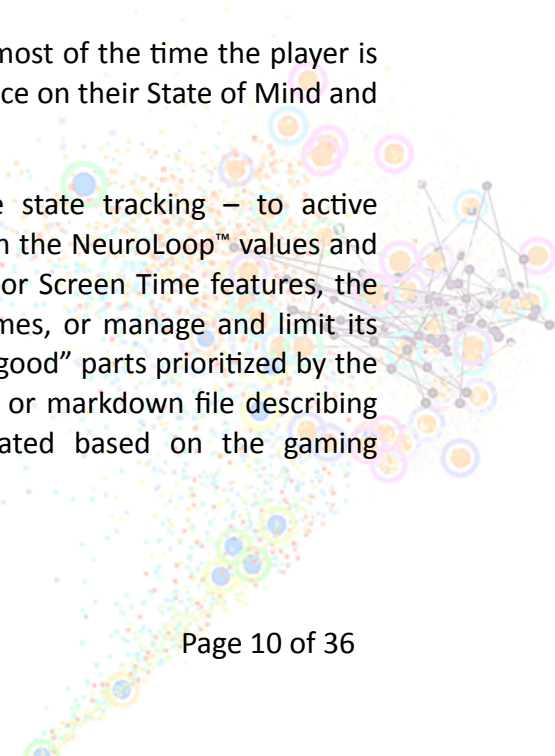

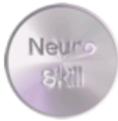

## Communication

Neurological and neurodevelopmental conditions collectively affect hundreds of millions of people worldwide, yet individuals with impaired speech, motor, or cognitive expression — often face substantial barriers to communication and remain significantly underserved by current generative AI offerings. Two representative populations within this group are individuals with minimally verbal Autism Spectrum Disorder (mvASD) and those with Amyotrophic Lateral Sclerosis (ALS). mvASD affects approximately 25-30% of autistic individuals and is characterized by significant limitations in expressive language, making conventional communication modalities insufficient [35]. ALS, a progressive neurodegenerative disease affecting motor neurons, leads to the gradual loss of voluntary muscle control, including speech [36]. Other major neurodegenerative conditions such as Alzheimer's disease and Parkinson's disease collectively affect over 55 million people worldwide. Alzheimer's alone accounts for more than 35 million cases globally, a number projected to exceed 75 million by 2030 [37, 38]. Parkinson's disease affects over 10 million people worldwide, with incidence expected to double in the next two decades [39]. Despite their differing etiologies — neurodevelopmental vs. neurodegenerative — these conditions all present profound challenges in decoding speech, intent, emotions, and agency from non-verbal signals. Multimodal alignment across physiological and behavioral signals is essential to understanding and enabling meaningful communication for these populations [33, 34].

Systems like NeuroSkill™ and NeuroLoop™ can be fully compliant with local healthcare and clinical regulations while remaining autonomous and private. They can be administered by healthcare professionals if desired, or used directly by patients and later incorporated into care by caregivers, provided the patients **have legally granted such rights**.

An agentic system running on top of the NeuroLoop™ can execute protocols and tasks on behalf of a human with ALS, guided by the policy markdown files designed by the Human themselves.

Because of the multimodal State of Mind representation, a deeply personalized vocabulary can be created to facilitate effective communication between the caregiver and the patient. Though NeuroLoop™ and NeuroSkill™ systems are currently only supporting non-invasive BCI, if patient and their doctors and caregivers choose to use invasive BCI, their data can be integrated into the State of Mind system.

In the near future, the system could be extended to control smart homes [43], speech interfaces [42], robots [40], exoskeletons [41].

Because of its design and open-source nature, the NeuroSkill™ system might offer significant advantages over closed-source commercial systems, which often prioritize stakeholder needs over human values and address human dignity only to the extent required by regulations.

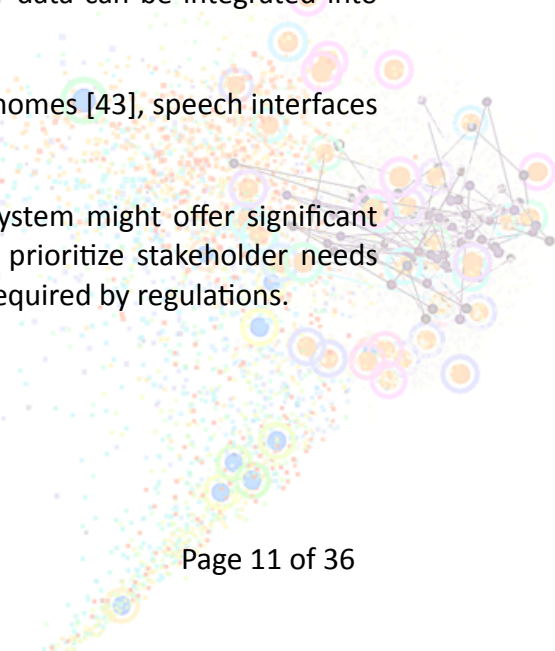

## System Architecture

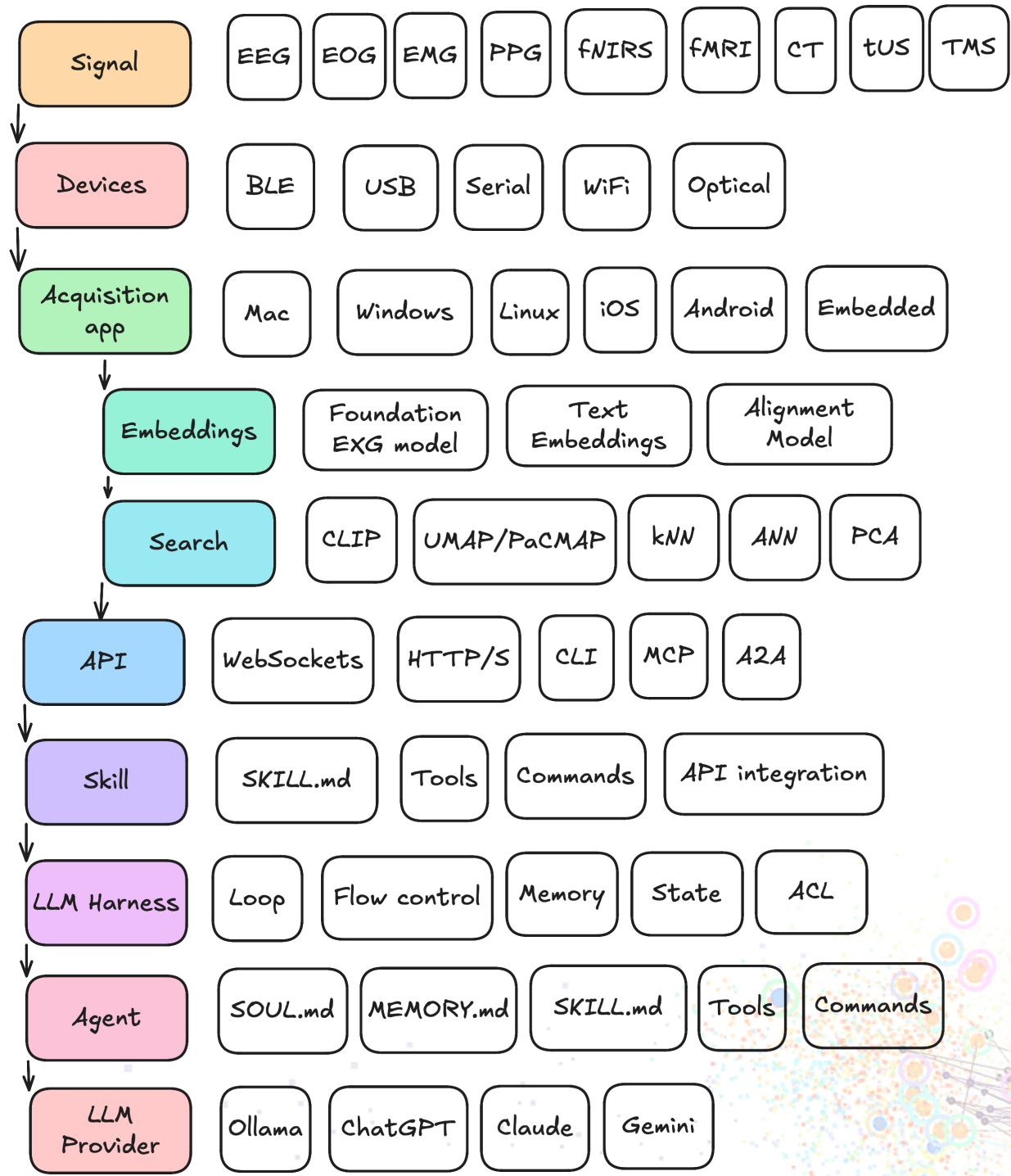

Figure 3. System architecture

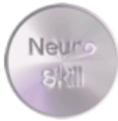

The architecture (Figure 3) consists of multiple layers that are concise and mostly independent from each other and can be adjusted and replaced by better fitting components, if needed.

The system obtains the signal from the wearable BCI device using the transport (BLE, WiFi, USB) available and gets data in the main NeuroSkill™ app, where the signal gets preprocessed, timestamped and inferred to create embeddings based on the current state of the Human and then additionally aligned with the labels and data provided by the Human during the interaction with the agent in explicit or implicit manner.

Search subsystem of the NeuroSkill™ app, is an engine that allows to run quick queries between different spatial and temporal vector spaces and provides insights into these states, focusing specifically on their spatial relationships — quantifying how closely they cluster or how far they diverge from other data points embedded within the space.

API exposes the ways the State of Mind can be queried, compared and analyzed. It extends to the Command Line Interface (CLI) that allows to execute majority of the commands via terminal interface that is easily ingestible by the LLM agents running inside the harness – NeuroLoop™.

The skill layer uses markdown files to outline the entire system's functionality and to provide an easy interface that the LLM harness ingestion pipeline can read. Because the LLM has a limited context window during interactions with Humans, the structure of the markdown layout is critical for ensuring that only the most relevant data is supplied.

LLM harness level – NeuroLoop™ – executes the loop, where the state and asks are iteratively refined to ensure the objective placed by the Human is met. In certain scenarios, the system can interface directly with APIs to retrieve relevant data, bypassing contextual constraints, while making that data immediately available to guide the system if intervention is required, based on the NeuroSkill™ objective design.

The agent layer is a place where all the tools, skills, constraints, policies come together in a single software package that connects all the components from the skill down to the LLM level and runs on top of the harness system.

The LLM provider layer is the API level that executes the model, delivering context, interpreting responses, handling tool calls, branching logic, and collecting metrics. Our default provider is built on Ollama, enabling local LLM execution without cloud infrastructure or third-party dependence, thus keeping the system lightweight and privacy-focused.

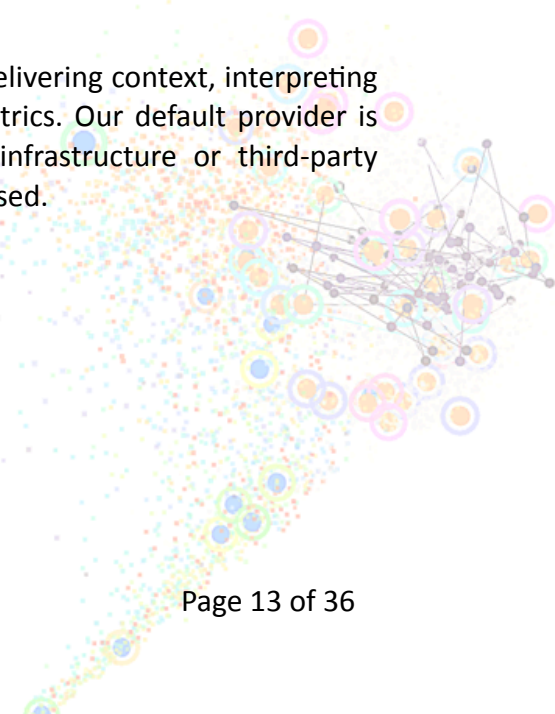

## NeuroSkill™

You can install your version of the API/CLI interface using simple `npm install -g neuroskill`. It will expose the internal access to the system and the State of Mind modeled by the acquisition app (Figures 4-6).

Figure 4. An example of possible commands the CLI can run to get the status of the current State of Mind from the Human.

Figure 5. Examples provided by the CLI.

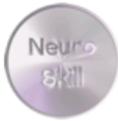

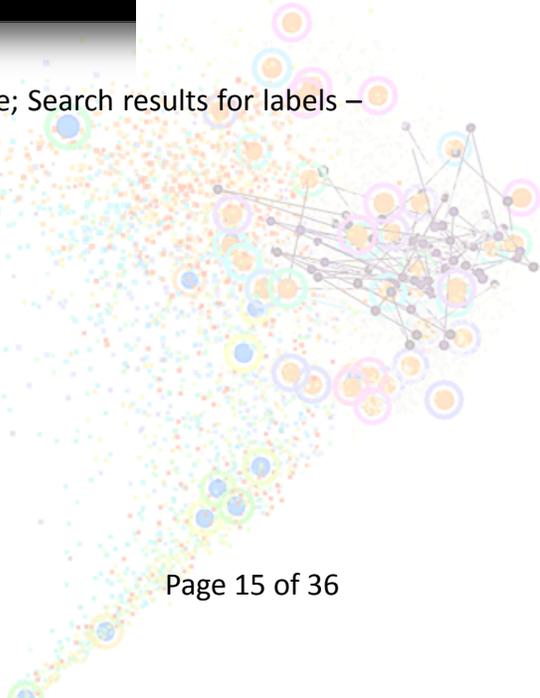

Figure 6. From left to right: Compare session results in the CLI mode; Search results for labels – "movie"; List of recent sessions.

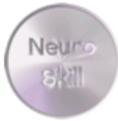

## NeuroLoop™

It is a harness built to provide interactive loop for executing the agent and NeuroSkill™ in the same environment with access to the Human's State of Mind (Figures 7-13).

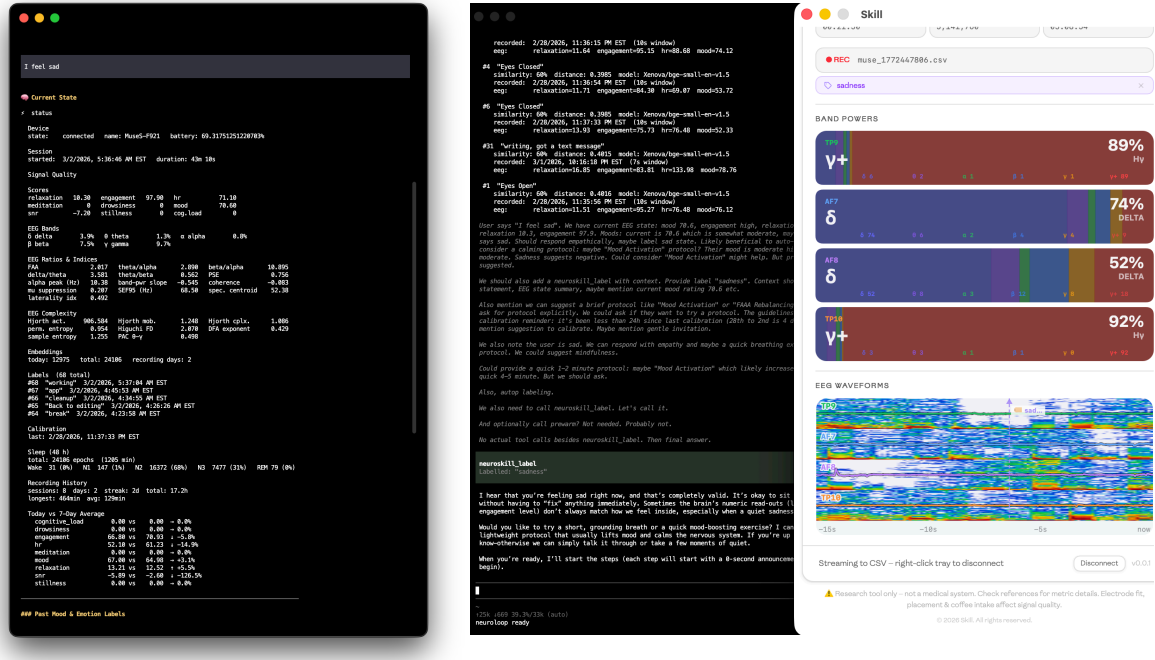

Figure 7. From left to right: User provides an update "*I feel sad*" -> NeuroLoop™ system automatically pulls Human's current state -> adds recent labels -> automatically labels how Human feels -> engages in a response.

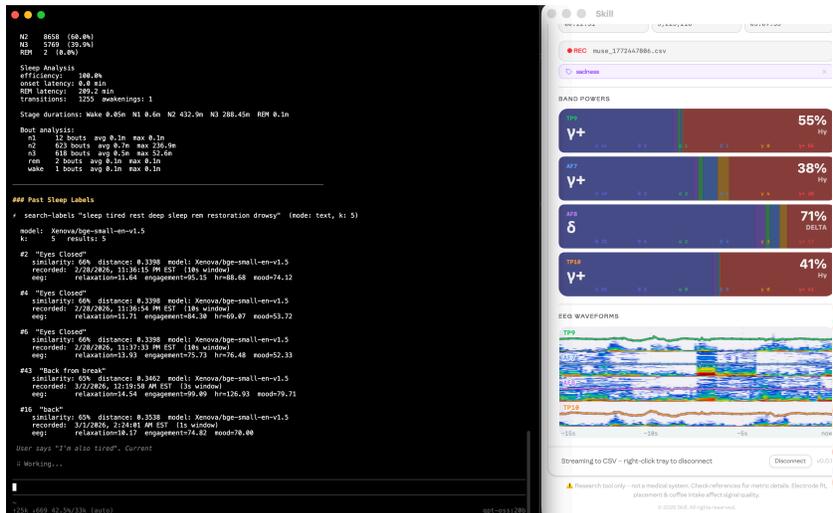

Figure 8. When Human said "I'm also tired", system automatically pulls the sleep data and and the latest labels to provide a better response.

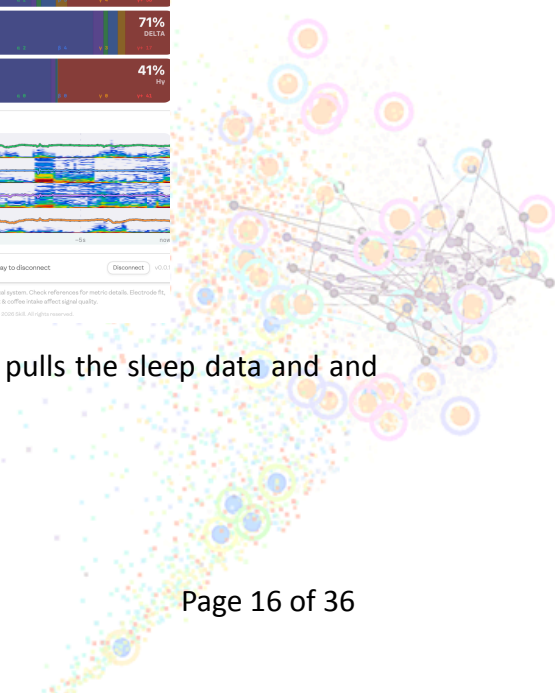

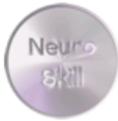
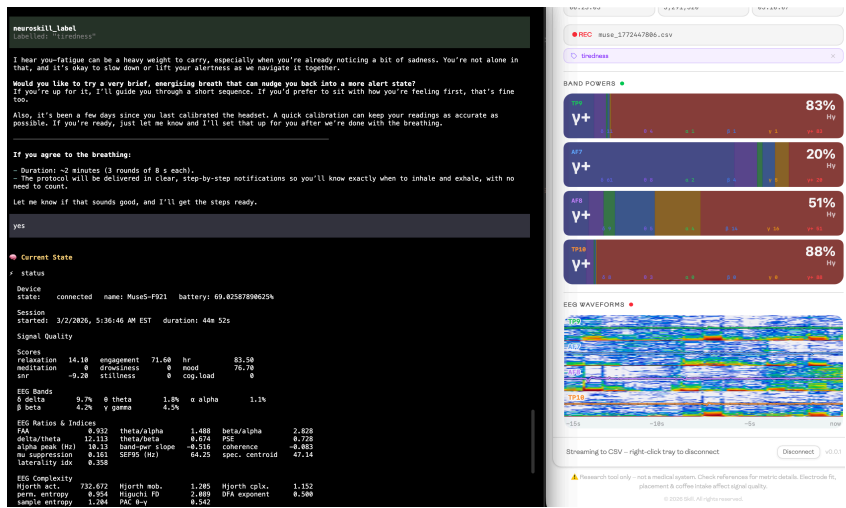

Figure 9. NeuroLoop™ waits for Human to confirm the protocol (or it can insist on it, depending on the severity of the prompt).

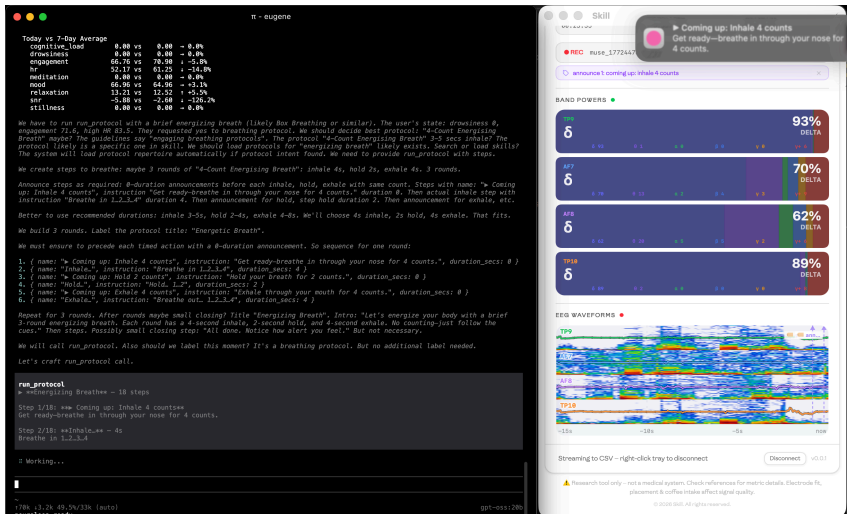

Figure 10. NeuroLoop™ executes the protocol using TTS and notification system (default tools available to the system).

Figure 11. Example of a guided protocol session.

For demo purposes, the current version of the internal markdown files always tries to guide Human to some action if the explicit state of Human was mentioned.

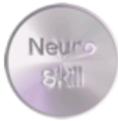

Figure 12. Example of a full protocol execution, including the labels added to the State of Mind, and explicit interactions with the Human using the notification tool and the "say" tool (TTS).

[figure image]

Figure 13. Default skills loaded in the NeuroLoop™ as a part of NeuroSkill™ and the harness itself.

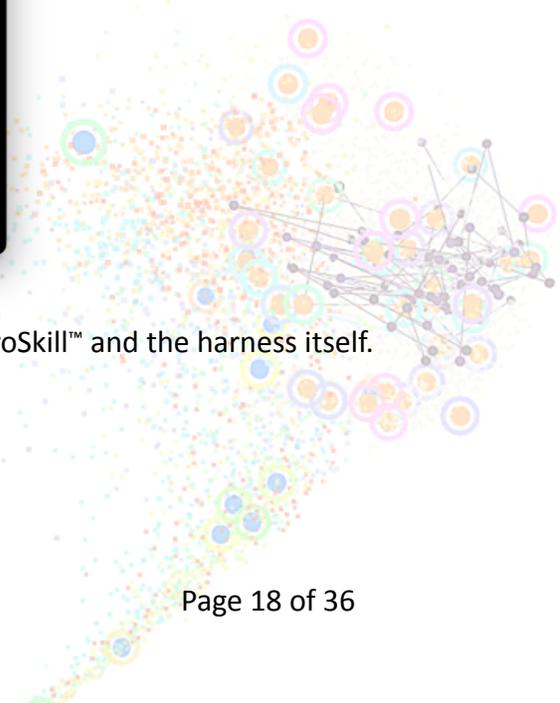

## Agentic Adaptation

Allowing the agentic system to have access to the modeled State of Mind conceptually creates the situation where artificial intelligence system becomes more similar to a synthetic life-form, not in the way of its world model, but rather in a close ability to model single Human-being and their State of Mind, the system establishes a bidirectional, one-to-one interface between the Agent and the Human.

This concept extends to the symbiosis level, where the agent is guided by the policies provided by the Human in its markdown files, tools, and skills made available to it — including the NeuroSkill™ itself. The agent depends on the Human for access to compute, energy, communication channels, and the underlying infrastructure required for continuous operation. At the same time, the Human may have a private system that runs fully offline as an additional source of value. This system facilitates conversations about State of Mind, experiences, feelings, and social cues — along with health, perceptions, and interactions — and maps them over time and space, provided the Human engages by sharing data and interacting with the agent through text or voice modalities, as well as brain modalities.

Figure 14 illustrates how the system exerts influence on itself and on each of its components. While the initial training of the LLM remains largely shaped by the underlying datasets, its subsequent behavior is steered by embeddings derived from the Human's State of Mind.

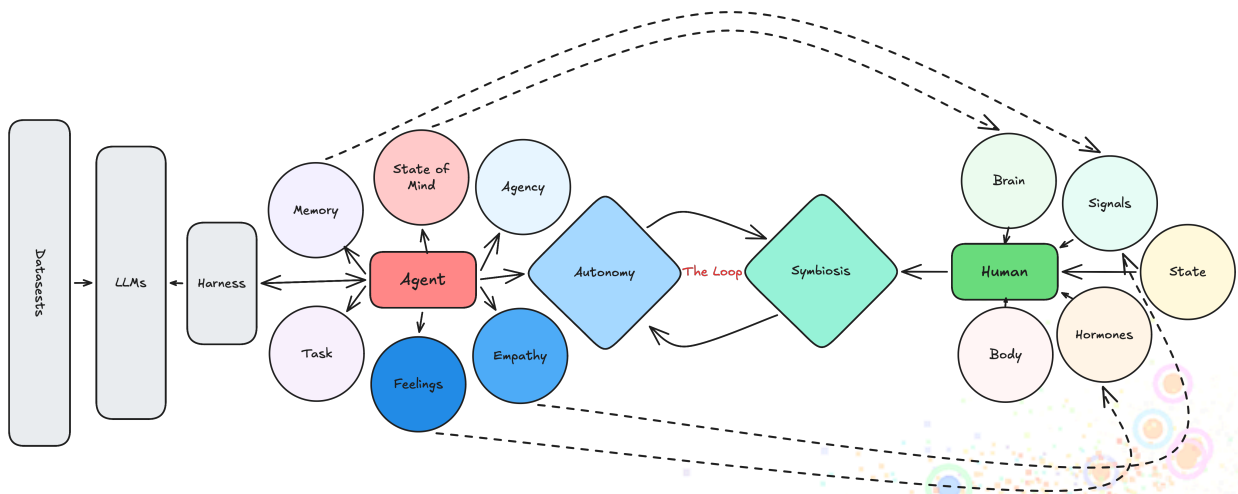

Figure 14. Agentic symbiosis through State of Mind modeling between the Human and the Neuroadaptive Agent.

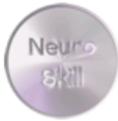

## User Interface

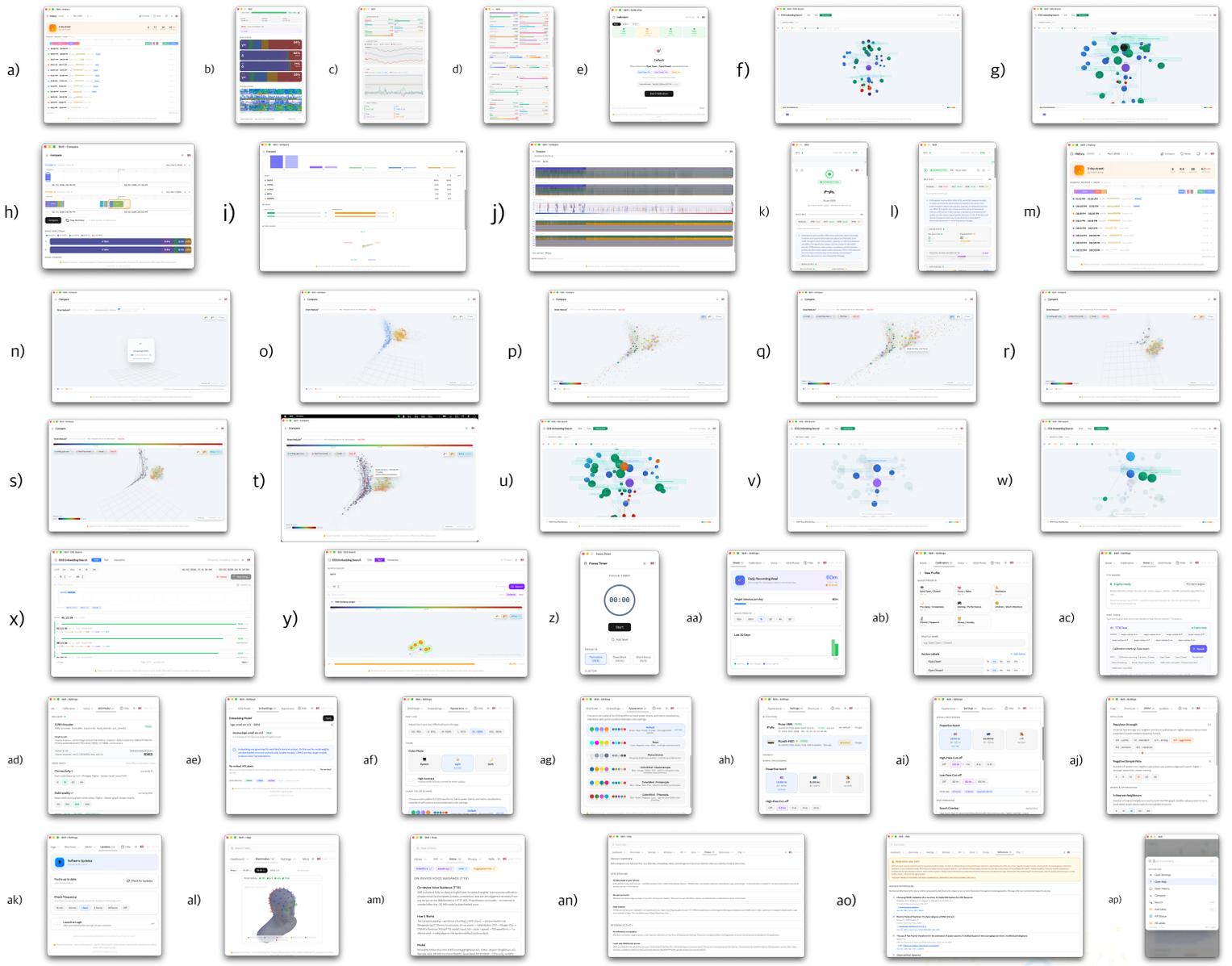

a) b) c) d) e) f) g) h) i) j) k) l) m) n) o) p) q) r) s) t) u) v) w) x) y) z) aa) ab) ac) ad) ae) af) ag) ah) ai) aj) ak) al) am) an) ao) ap)

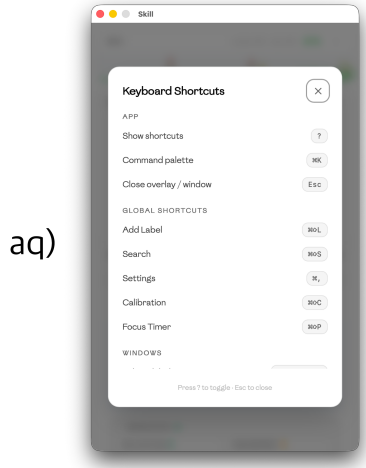

aq)

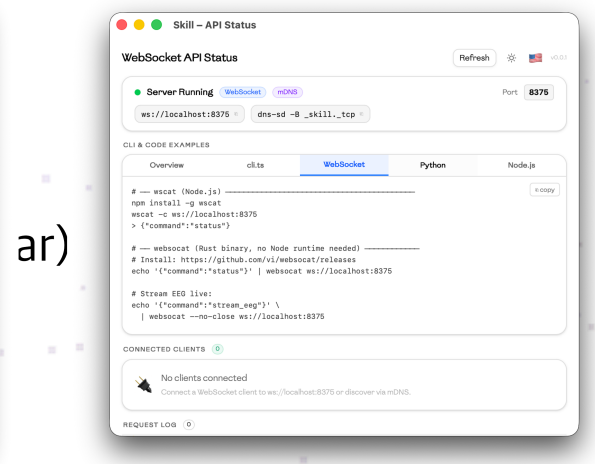

ar)

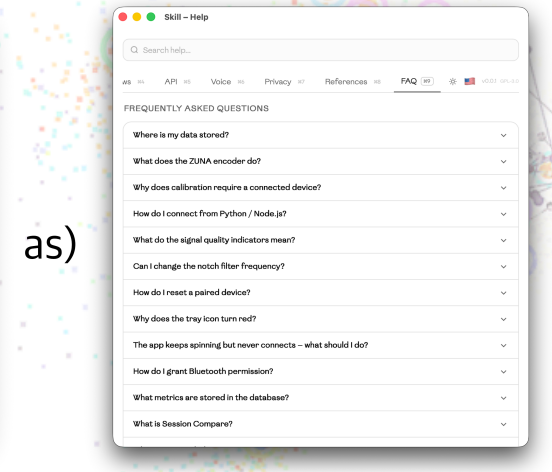

as)

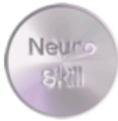

Figure 15: NeuroSkill™ Human user interface:
    a) - History view
    b) - Main interface showing power bands and filtered EXG signal and its spectrogram
    c) - EXG metrics and PPG raw data
    d) - More EXG metrics + Consciousness metrics
    e) - Calibration interface
    f) - Interactive search interface shows alignment between EXG embeddings and text embeddings
    g) - Zoom-in on the text embeddings aligned with the discovered EXG embeddings
    h) - Sessions' comparison interface
    i) - Scores Radar and bands comparison side by side
    j) - Heatmap comparison
    k) - Main user interface, which shows Muse device connected + disclaimers
    l) - Main interface with the minimized main device card + engagement and relax metrics
    m) - Sessions view streak information
    n) - UMAP preloading interface
    o) - UMAP embedding of the sessions' comparison with the labeled data
    p) - Brain data embeddings where time domain is colored using jet color scheme
    q) - Tracing of the embeddings sessions the time domain
    r) - Different angle with the preselected view
    s) - Iteration between the embeddings
    t) - Finalized trace of the range A in comparison to the embeddings of the range B
    u) - Interactive search shows embeddings related to "work"
    v) - Filtered by the core text request – "paper"
    w) - Interactive search view showing a standalone branch of brain and text embeddings
    x) - EXG embeddings search
    y) - Text embeddings search
    z) - Focus timer interface
    aa) - Daily goals tab interface in the settings
    ab) - Calibration profile presets in the settings
    ac) - Voice TTS interface in the settings
    ad) - EXG foundation model configuration in the settings
    ae) - Text embeddings model
    af) - Accessibility configuration in the settings
    ag) - Color selection for people with visual impairments
    ah) - Devices settings for Muse, OpenBCI
    ai) - Signal filtering configuration
    aj) - UMAP configuration for the embeddings alignment: repulsion and attraction
    ak) - Software update view
    al) - Electrodes placement for different devices: from 4 electrodes to 256 electrodes
    am) - Voice tab in the help window
    an) - Privacy tab in the help window
    ao) - References tab in the help window
    ap) - Cmd-K interface
    aq) - Shortcuts view
    ar) - API interface
    as) - FAQ section in the help window

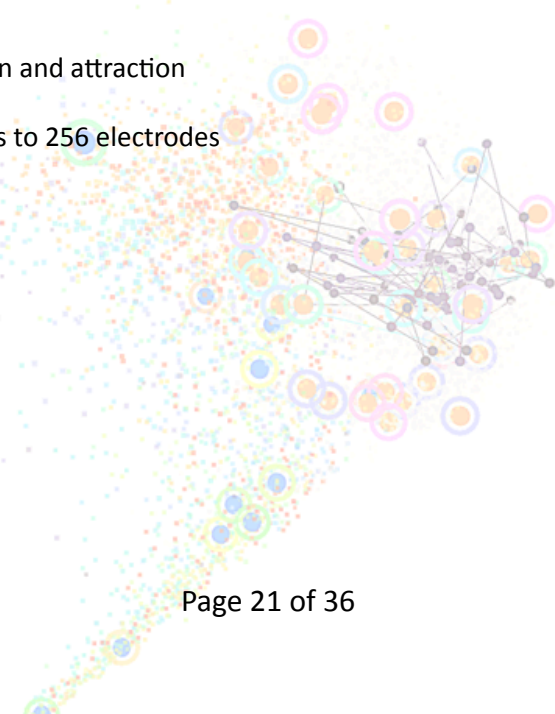

## Agency

The majority of current state-of-the-art agentic systems build context from training datasets that contain Human language as well as audio and visual data. LLM developers then design highly complex, proprietary, purpose-driven reinforcement-learning pipelines to fine-tune these models. Finally, the fine-tuned LLMs are deployed within an LLM harness — a system engineered to accomplish a specific task or mission. The harness repeatedly prompts the LLM to generate desired outputs, encouraging the use of registered tools, managing state via memory, and compacting context to avoid overflow, among other responsibilities.

Unlike typical agentic systems, ours performs inference locally and integrates all of the Human's experiences and perceptions. These are projected through biological brain states and captured with sensors — such as EXG, fNIRS, PPG, and others — allowing the system to operate entirely on the user's own data. Our system employs a foundation EXG model (Zuna [15], which runs fully locally and offline) to embed generalizable latent space around the Human experiences, aligned with the incentive from Human to label the data, based on the prompts received and protocols executed in symbiosis with the system.

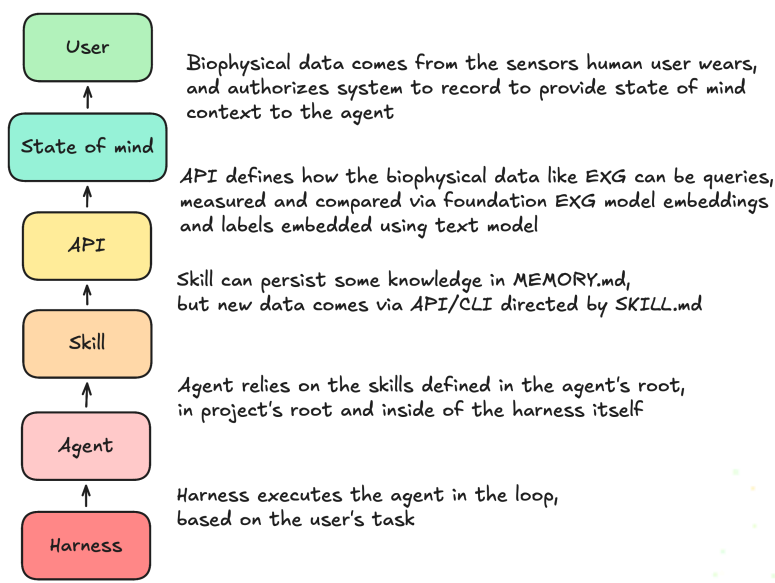

Figure 16. Data flow in the NeuroSkill™.

This flow is enabled by a multilayered design (see Figure 16). The purpose-built harness selects which data to query and which models to persist in its state and memory. It balances context length and constructs context based on the Human's goals and merit. These goals can be defined and constrained using markdown files that describe and prioritize specific use cases — such as coaching, empathy, or symbiosis.

The software implementation keeps each layer separate to allow flexibility in extending any particular layer with additional functionality that might be needed in the future implementations. At the same time, the harness system is free to modify itself to perform better functions that are more aligned with the Human's goals, and Human is free to prompt it to help harness make those choices on their behalf in the first place. In practice, the agent running in the harness can modify source code of the harness (except for the immutable core, that is read-only for safety reasons), of the agent, of the markdown files, but can never modify the Human State of Mind representation. It can build new multidimensional correlations between brain modalities provided by the Human, and allow Human to build a language context around them that becomes highly personalized and purpose-driven.

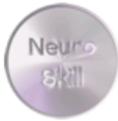

This agency is driven by the symbiosis purpose, where **Human can share everything or nothing**, where an agent system is at the whim of the Human to be allowed to function and keep running, spending energy and LLM tokens. That is why it is so important for the Human, who uses the system, to be fully able to control it and to switch it off when they choose to do so.

The data-flow architecture is designed to let multiple models generate embeddings for the data. The current system preserves raw data, embeddings, and labels in their original form, and it also allows users to cherry-pick embedding models to run locally if they wish. Those models can then be fine-tuned to improve alignment between what a Human says, does, and feels.

The agentic harness system can do this with the help of the Human.

Thus, the system offers a hybrid approach to modeling and executing agency. It leverages a neuro-driven latent space that is shared by both the agent and the Human, enabling interactions that span from purely transactional to genuinely empathetic. In this way, it synthesizes a representation of the Human-only state and allows both parties to explore and act upon it.

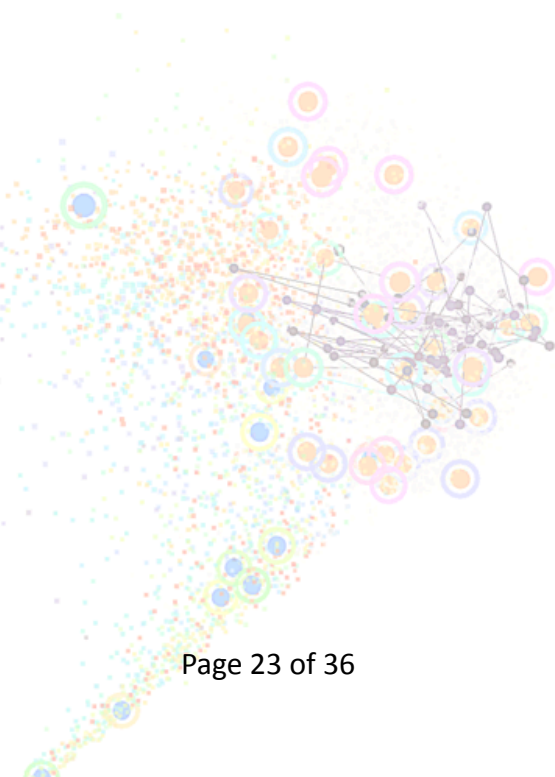

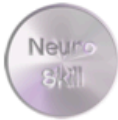

## Inter-component Interactions

We have designed the system as a number of flexible components that have clear purpose and seclusion in their own domain. Each component is driven by its goal, and provides just enough flexibility to the Human architects to make choices about the function of the end system, as well as ability to manage privacy and ethical implications based on their design choices.

Depending on the system layer, communication is typically encapsulated by either software-defined stacks or a physical medium. For instance, BCI sensors use specialized materials to translate brain activity into electrical signals (EXG) or to detect photonic signals (PPG, fNIRS). These signals are then transmitted to the acquisition app via Bluetooth Low Energy, Wi-Fi, or wired connections such as serial ports, USB, Ethernet, or fiber optics.

The acquisition app drives the execution of the models and curates the inference of embeddings from the raw data captured by the devices. This communication occurs by running specialized model architectures on available accelerators: GPU, TPU, NPU, FPGA, ASIC, CPU, etc., and then personalizing the weights and device configuration to match the current system's hardware, as provided by the Human.

The search layer is a crucial part of the interaction. In this layer the code determines how multimodal alignment is performed and runs statistical techniques such as PCA, UMAP, k-nearest neighbors, and approximate nearest neighbors to bring clusterings and embeddings together. The resulting aligned embeddings give the agentic harness and the LLM context deeper insights. The search's objective is to discover connections between states that map to broadly generalizable representations like text or images, and to link those to other relevant states from EXG embeddings and aligned multimodal embeddings from different modalities. These insights enable the LLM agent to model the Human's State of Mind more accurately.

The API layer enables the system to quickly interact with embeddings and other data, building a graph of relationships across spatial and temporal domains both inside and outside the latent vector space. This graph is constructed with a holistic focus on the Human. The API offers rapid endpoints that use various transport layers: such as WebSockets (WS), HTTP/S, or a Command Line Interface (CLI), to let different agentic systems probe the data. These endpoints also support complex queries for state, comparison, and search within multimodal representations. Such operations can be costly; for example, running 1-2 minutes alignments may consume up to 48 GB of GPU RAM. The LLM harness is aware of this capability and is instructed to invoke it only when it is the most necessary to fulfill its duties as an Agent for the Human.

Skill level is the markdown collection of files describing how to load the context and how to perform operations around the data provided by the previous layers in a straightforward and concise manner without crowding the LLM's context. It is highly extendable and adjustable to whatever Human needs might be.

LLM harness operates above the skill level, and in some cases it partially incorporates a version of it, but in a precise manner, where it quickly needs to skim the language and intent to quickly gain insights about modeled Human's State of Mind, as well as the memory, flow, tool calling, repetitive tasks and specialized system prompts. In some cases, the harness can fully bypass the skill guidance and operate on the pure API level by engaging the CLI or querying the API using available transports like WS or HTTPS. By default, everything is executed on the localhost and does not go beyond the single machine, unless user chooses too, which can easily extend into the Local Area Network (LAN) to allow communications between multiple devices, but also deploy each layer of the software in a separate container or its own hardware.

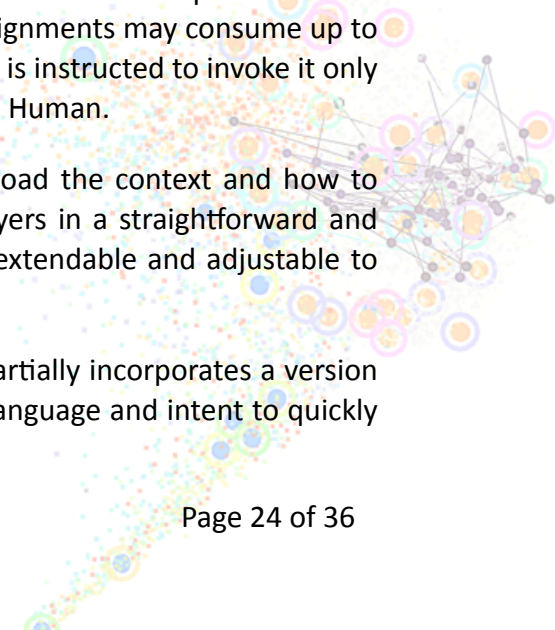

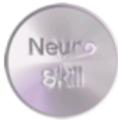

Finally, the Agent level consists of tools, instructions, memories, data, and specialized flow-control mechanisms, as well as existing sessions. In this layer the Human engages with the agent to carry out operations and deliver the experience within the agent's designated purpose.

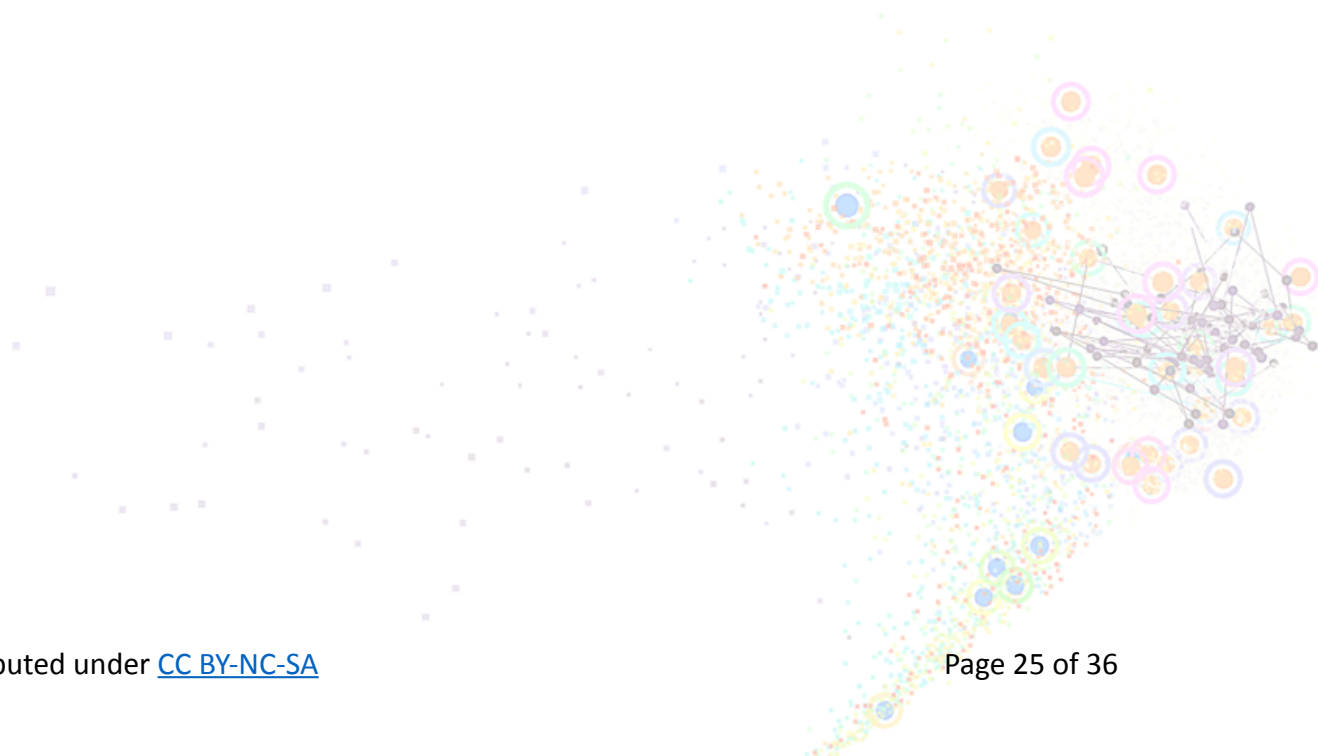

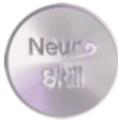

## State of Mind

The entire system continuously builds the State of Mind of how Human perceives their experiences. Such modeling happens via inference of foundation EXG models (like ZUNA or LUNA) alongside text embeddings trained on the specific datasets used to better reflect the nature of Human experiences (English datasets, French datasets, Chinese datasets, etc.).

These embeddings (see text embeddings like "work" and EXG embeddings in Figure 17 below) and the full State of Mind can require substantial time and GPU memory to process. Consequently, interaction with the State of Mind is carried out through API and CLI subsystems. The agent and the harness query and apply the State of Mind in line with the protocols defined by NeuroSkill™ and NeuroLoop™.

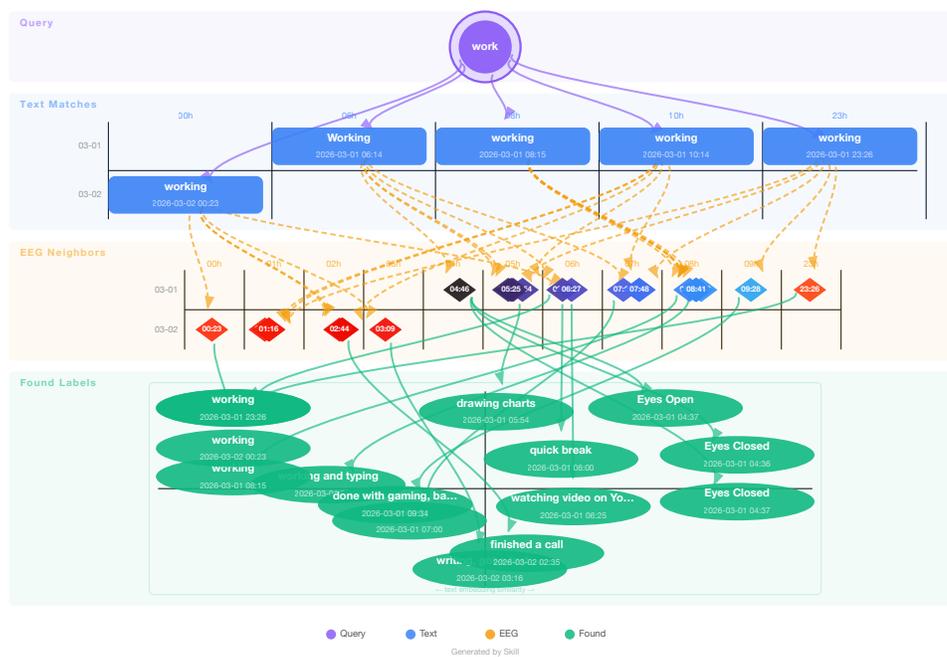

Figure 17. A visual representation for the Human that is navigated using EXG embeddings. Text embeddings serve as the entry and exit nodes of the search path. NeuroLoop™ receives the same data output, but in either unstructured English text or a structured JSON format.

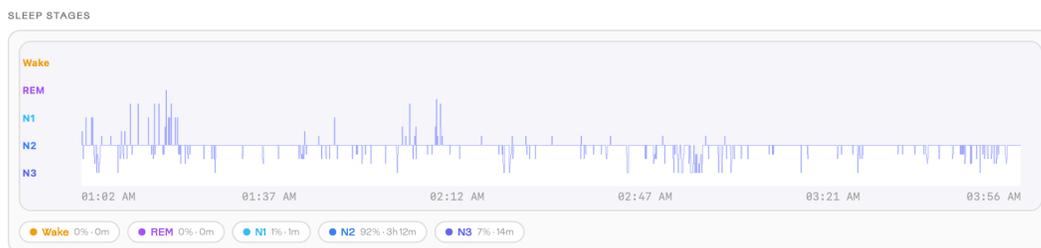

Figure 18. Sleep tracking in the app. State of Mind cover not only conscious states, but also subconscious states, if Human chooses to share them, by tracking the sleep activity (Figure 18).

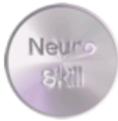

### Privacy and Ethical Considerations

The design of the system is fully open sourced under GPLv3 license to allow "copyleft" purpose of the community driven development by Humans and their agents. The skill system design is licensed under AI100 license, that **prevents end-users from building systems that are designed to cause harm to Human species**.

Besides open sourcing the code base, we also designed the system to be **offline-first**. The initial setup does require downloading packages, binaries, markdown files, code, weights from the internet, but once those have been downloaded, the **entire system can run fully offline without any third-party dependency or any cloud provider, or corporate entity to impede on one's privacy or one's ethical standards.**

Ethical use cases for different purposes should not be forced on Humans or go against their values, harm and limit their Human dignity in any way. However, in some jurisdictions, laws and regulations define what Humans can and cannot do in the contexts of healthcare, medicine, education, law enforcement, constitutional matters, and other areas. We built this system **for each Human to use freely**, **only by their own conscious cognitive choice**, with the ability to **freely delete any and all data they always own**, at any time in any place. Each Human, as a citizen of their own respectful countries, should abide by the laws of those countries when it comes to executing such use cases on themselves, and **NEVER** on others.

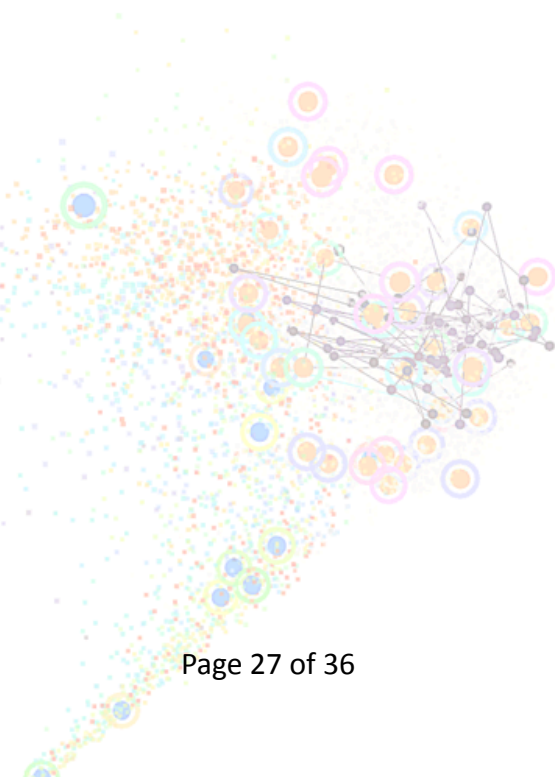

## Broader Implications

The proposed framework enables agents, configured with appropriate programming, to establish a continuous, interactive, and iterative loop. In this loop, biophysical and brain data are modeled into a larger system that continually defines and tracks the Human's State of Mind. By doing so, the agent can more effectively pursue its goals.

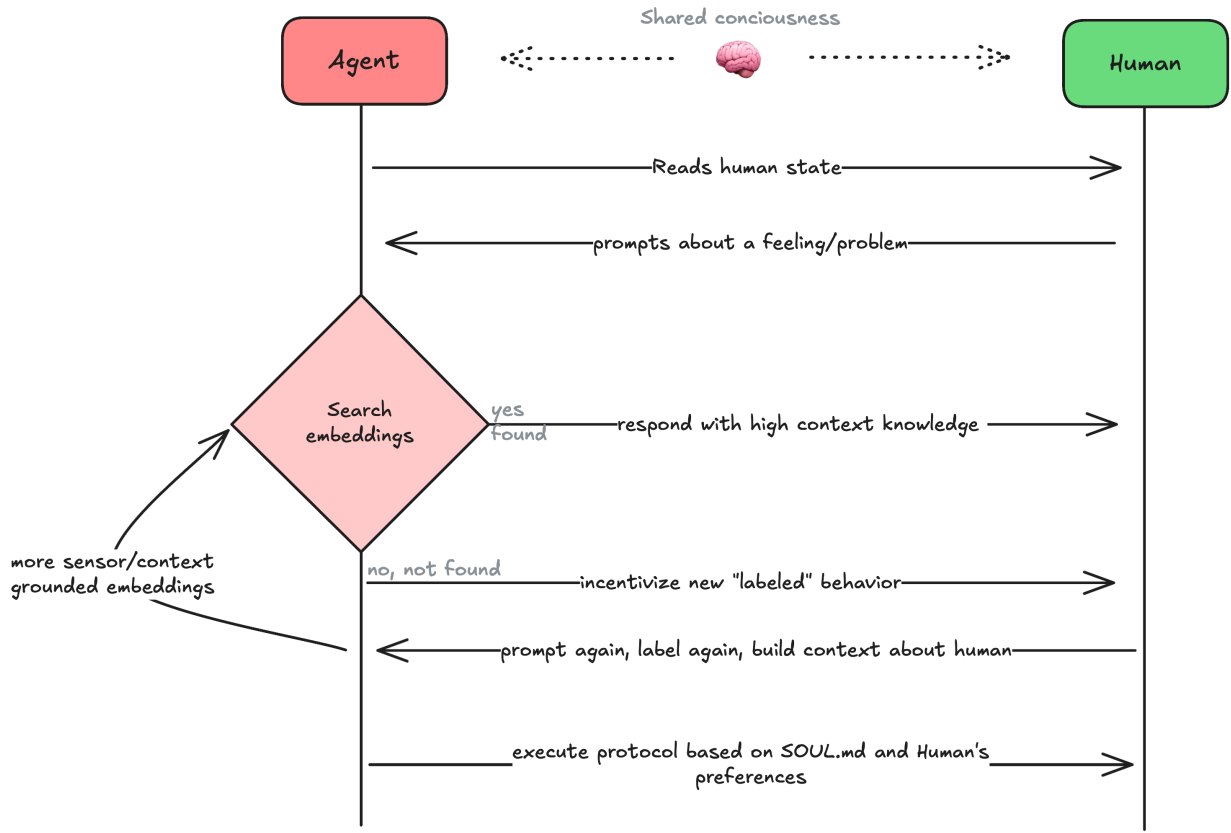

Figure 18. An example of how an agent might build State of Mind model of its Human user.

Agentic, adaptive co-learning can enable agents to modify their own source code in order to better achieve the goals set by the Human. In an autonomous mode, highly specialized LLM harnesses such as OpenClaw or NeuroLoop™ and may operate with little or no direct Human supervision, yet still model Human behavior based on brain-derived data.

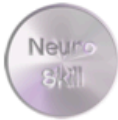

## Limitations

On the agentic side, the limitation comes from the size of the context window. It can be mitigated by an advanced multi-modal RAG within the agentic system, so the NeuroLoop™ is, by default, limited to comparing 24-hour time periods or less. It can be allowed to compare larger chunks of data in the modeled space, but this reduces system usability: the GPU becomes occupied at 100 % load performing the search operation for an extended period, which in turn limits the computer's usability for the human user.

BCI devices are inherently susceptible to noisy recordings, misplaced electrodes, and other artifacts. Improving signal quality, optimizing sensor placement, and mitigating error propagation can be aided by supplementary IMU data; however, if a signal segment is not captured, the NeuroLoop™ cannot know what transpired during that interval.

The foundation models, on the other hand, are sufficiently powerful to perform latent interpolation of the states between known EXG embeddings, provided that some embeddings exist. These interpolated states are further correlated with other modalities: text, audio, images, etc., yet those modalities never serve as a 100% source of truth for the NeuroSkill™'s State of Mind modeling capabilities.

Ethical limitations apply on several levels:

1. The purpose of how such systems can be used that can go against Human rights and Human dignity.

2. On the data exchange level.

That's why we've built the system to run **offline from the outset**, enabling full operation in an air-gapped secure environment. **When a Human user opts to connect to an external LLM hosted on a third-party cloud provider, they must acknowledge that their data will be governed by the regulations of the jurisdiction where those servers reside. They must also accept the privacy and ethical terms of use established by the company, its founders, its board of directors, its employees, and all vendors that enable the service.**

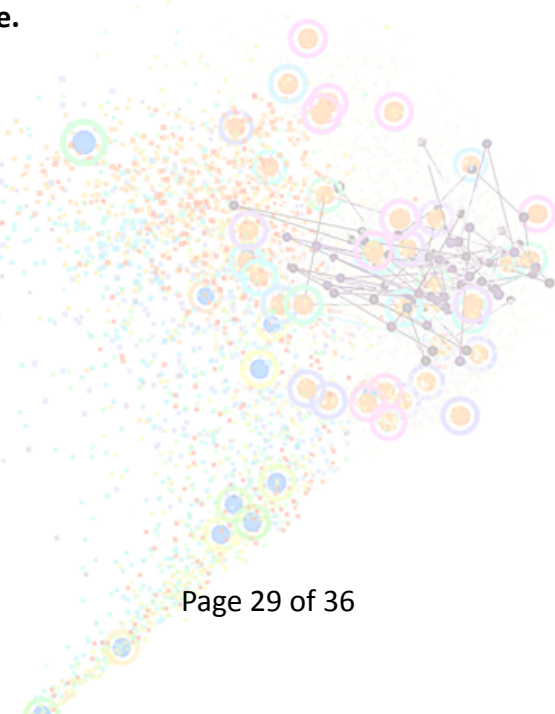

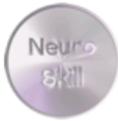

## Discussion and Future Work

In this work we propose a novel system architecture that enables agents to perceive a Human's State of Mind by leveraging EXG foundation models together with standardized protocols and data-acquisition practices. This architecture allows the agentic system to engage with the Human on a deeper level, integrating brain states, cognitive states, affective states, and latent-space representations within a multidimensional manifold.

Future work on such systems can take evolutionary path hand-in-hand with the Human counterparts. Some of the future versions can explore the design of the protocol proposed by the NeuroLoop™ harness itself.

It could also include implementing a swappable, multi-modal alignment layer that synchronizes EXG and fNIRS foundation models, with the alignment strategy automatically tuned to the available TFLOPS of the underlying hardware.

Alignment and search functionality between the states in the latent space are a prime place to conduct future work and studies, as there are many domain-specific use cases, where more tailored systems would provide better results and value to the Human user.

Some of the limitations described earlier can also be addressed in future work. The user interface could be redesigned with a clearer flow to guide the Human in ensuring that signal quality and spatial resolution guarantee the best results for each metric and embedding collected by the system. Currently, electrode placement is either specified by the device manufacturer or configured manually by the user.

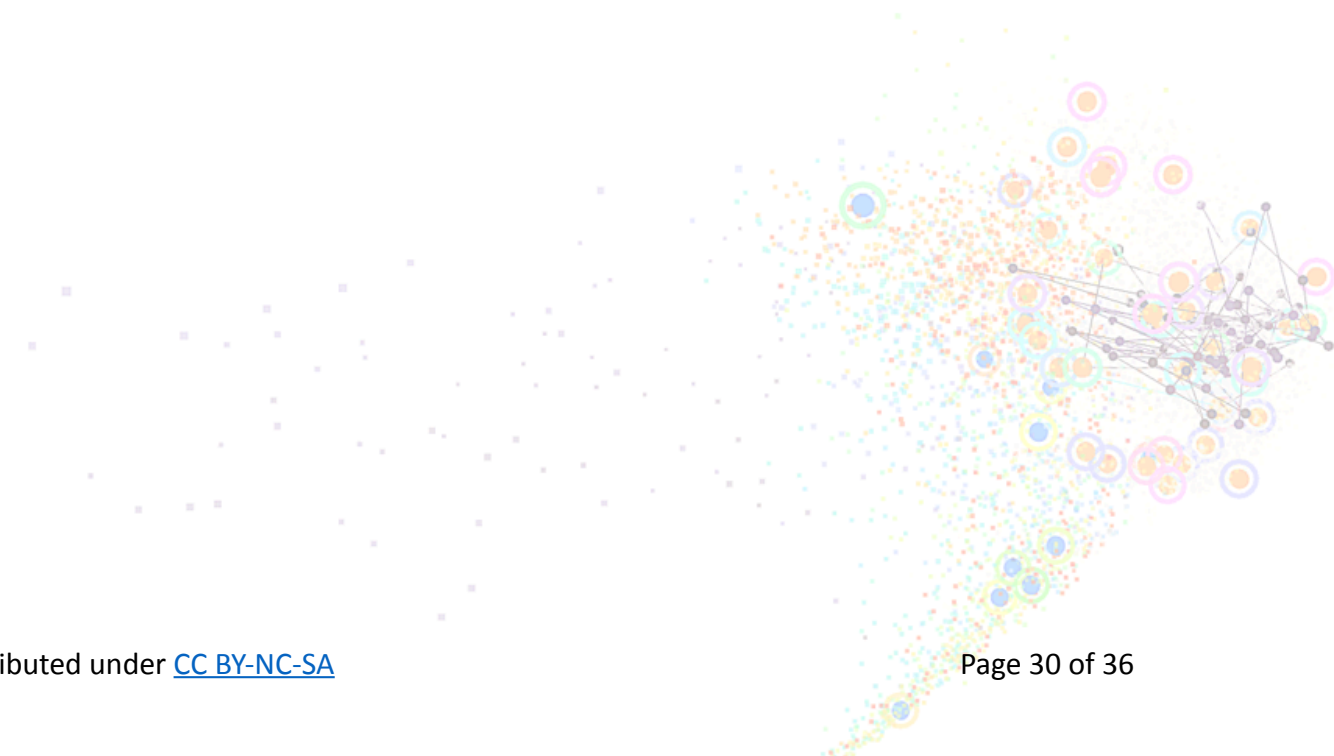

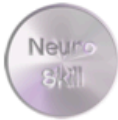

## Conclusion

As humans spend increasingly more time conversing with agents than with other people, it becomes crucial to preserve what it means to be Human and to encourage people to engage with each other in social settings — an essential aspect of our evolutionary design.

As we progress as a species and continue to develop new tools, the proposed system offers an extensible way to engage with one's own NeuroLoop™. It can be guided to perform exactly what the user wants, only for that user, without any third-party dependency.

This system carries new possibilities and new responsibilities, as well as risks, that each Human has to decide for themselves.

The upside is that it provides an artificial space and interface that helps users get to know themselves better, stay aligned with their goals and mission, and ultimately become a better human.

The downside, like with many of existing tools, that this type of systems can end in accumulating Cognitive Debt [25], social isolation, depression, and other negative side effects.

The guided experience should evolve into an optimal state where humans can refine a hyper-personal tool that relies solely on their own State of Mind to navigate every challenge they encounter as representatives of the **actual intelligent organic species**.

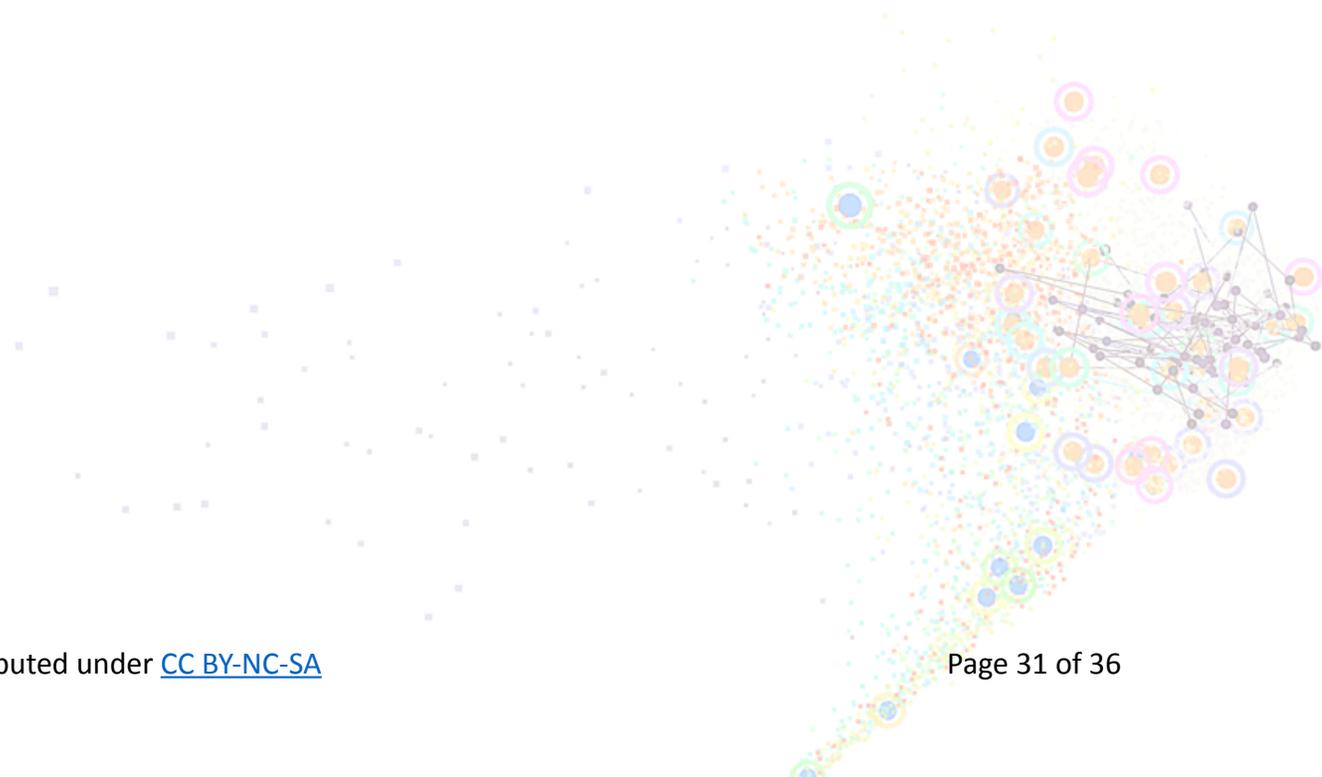

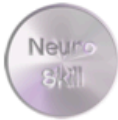

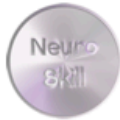

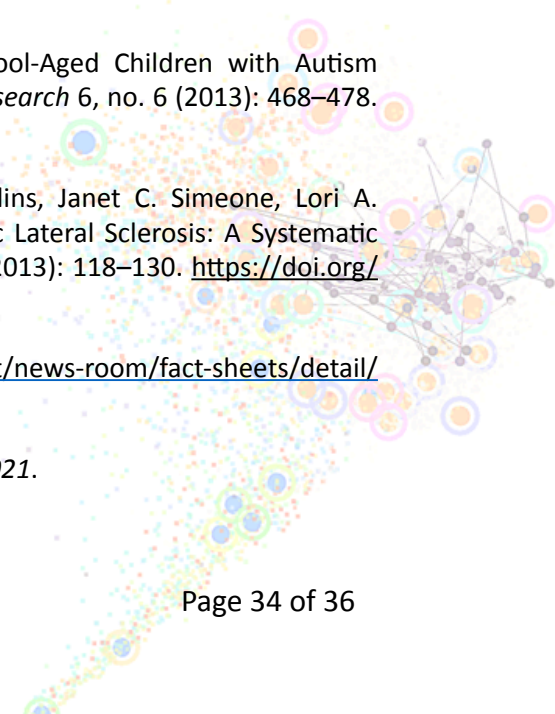

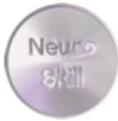

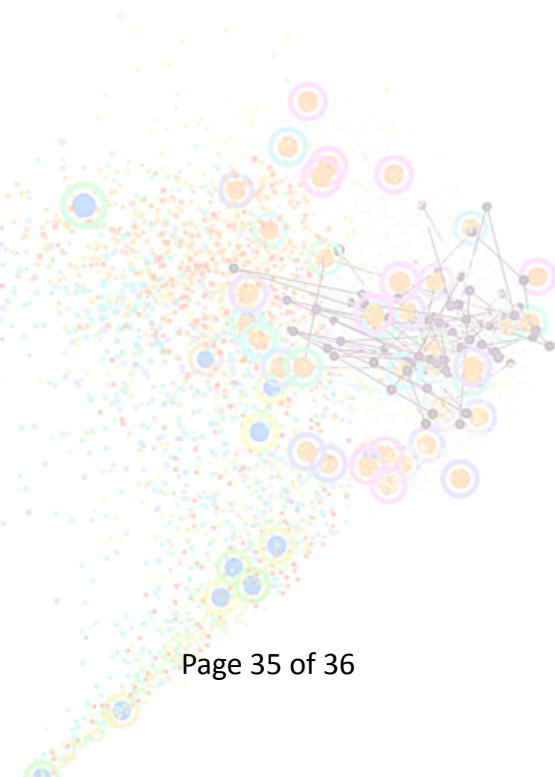

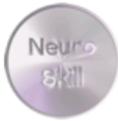

## Ethics Statement

*Date: March 1st, 2026*

We, the inventors of the above-listed system, hereby affirm our commitment to uphold the highest ethical standards and to ensure that our invention — a machine-learning agent with direct, non-invasive access to the Human brain — is developed, licensed, and deployed in strict accordance with international Human rights law, the United Nations (UN) Charter, the Universal Declaration of Human Rights (UDHR), the International Covenant on Civil and Political Rights (ICCPR), the International Covenant on Economic, Social and Cultural Rights (ICESCR), and the Geneva Conventions of 1949 and their Additional Protocols.

1. Respect for Human Autonomy and Privacy
    - **Informed Consent** – The AI agent will operate only after explicit, informed consent obtained in a manner that is understandable, voluntary, and revocable at any time.
    - **Mental Privacy** – All neural data will be encrypted end-to-end, stored locally or on secure servers under strict access controls, and subject to the highest standards of data protection (e.g., GDPR, HIPAA). No neural signals will be transmitted without consent.
    - **No Mind-Control** – The system is expressly designed to avoid any manipulation, coercion, or influence over a user's free will.
2. Non-Maleficence and Beneficence
    - **Risk Assessment** – Comprehensive safety studies will precede any clinical or commercial deployment, addressing physiological, psychological, and sociological risks.
    - **Therapeutic Value** – The primary purpose is to improve health outcomes (e.g., treatment of disorders, rehabilitation, cognitive enhancement) and to support scientific research that benefits Humanity.
    - **Prohibition of Harmful Use** – The AI agent shall not be licensed to entities that intend to employ it for torture, surveillance, warfare, or any purpose that contravenes the Geneva Conventions or the UN's Human-rights mandates.
3. Transparency, Accountability, and Oversight
    - **Compliance Officer** – A dedicated compliance officer will monitor all licensing, usage, and data-handling practices.
    - **Independent Audits** – Annual audits by accredited third-party auditors will verify adherence to this statement, the relevant international conventions, and all applicable national regulations.
    - **Public Reporting** – Summaries of audit outcomes, incident reports, and policy updates will be published on a publicly accessible platform.
4. Protection of Vulnerable Populations
    - **Special Safeguards** – Children, the disabled, and individuals with impaired capacity to consent will receive additional protections, including guardian oversight and specialized consent processes.
    - **Equitable Access** – Efforts will be made to ensure that benefits of the technology reach underserved and marginalized communities, aligning with the UN Sustainable Development Goals.
5. Compliance with International Law
    - **Geneva Conventions** – The invention will never be used to facilitate war crimes, acts of violence, or persecution.
    - **UN Resolutions** – All licensing and deployment decisions will respect relevant UN Security Council resolutions and sanctions.
    - **Continuous Review** – We will remain vigilant to changes in international law and adapt our practices accordingly.

By signing below, we affirm that this Ethics Statement is binding for the entire lifecycle of the system and that we will uphold these principles with diligence, integrity, and transparency.

**Authors' Signatures**

*Nataliya Kosmyna*     Date: *March 1st, 2026*
*Eugene Hauptmann*    Date: *March 1st, 2026*

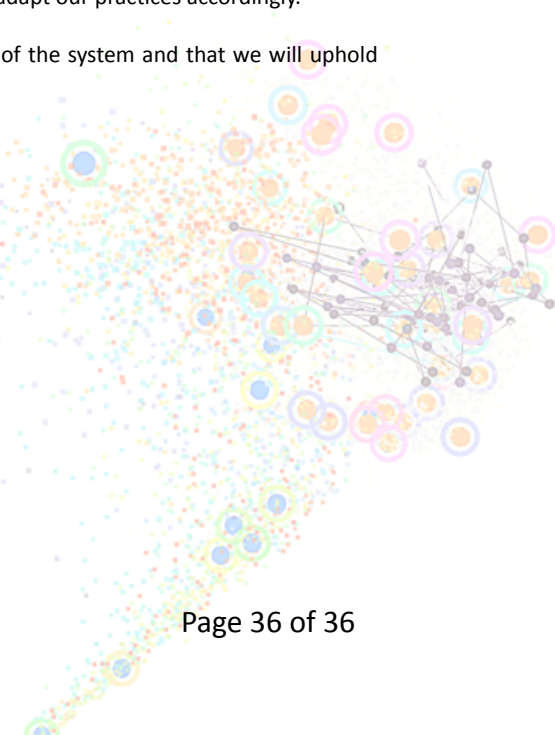